\documentclass{article}

\PassOptionsToPackage{numbers}{natbib}

\usepackage{algorithm}

 \usepackage[main, final]{neurips_2026}

\usepackage[utf8]{inputenc} 
\usepackage[T1]{fontenc}    
\usepackage{hyperref}       
\usepackage{url}            
\usepackage{booktabs}       
\usepackage{amsfonts}       
\usepackage{nicefrac}       
\usepackage{microtype}      
\usepackage{xcolor}         
\usepackage{graphicx}
\usepackage{wrapfig}
\usepackage{caption}
\usepackage{tcolorbox}
\usepackage{subcaption}
\usepackage{amsmath}
\usepackage{multirow}
\usepackage{enumitem}

\usepackage{algorithm}
\usepackage{algpseudocode}

\definecolor{best}{RGB}{220, 20, 60}     
\definecolor{second}{RGB}{30, 144, 255}  

\title{Harnessing Image Diffusion Prior for\\ Photo-Realistic Video Restoration}

\author{%
  Fanghua Yu$^{1,2}$,
  Hongyu An$^{3}$,
  Jinfan Hu$^{4}$,
  Xinqi Lin$^{5}$,
  Zhiyuan You$^{4,5}$, \\
  \bfseries
  Jian Wang$^{6}$,
  Hongyang Li$^{1}$,
  Chao Dong$^{4,7,\dagger}$,
  Jinjin Gu$^{3,\dagger}$ \\[0.5em]
  \normalfont
  \footnotesize
  $^{1}$~The University of Hong Kong \quad
  \footnotesize
  $^{2}$~Shenzhen Loop Area Institute \quad
  \footnotesize
  $^{3}$~INSAIT, Sofia University ``St. Kliment Ohridski'' \\
  \footnotesize
  $^{4}$~Shenzhen Institutes of Advanced Technology, Chinese Academy of Sciences \\
  \footnotesize
  $^{5}$~The Chinese University of Hong Kong \quad
  \footnotesize
  $^{6}$~Snap Inc. \quad
  \footnotesize
  $^{7}$~Shenzhen University of Advanced Technology \\[0.3em]
  \footnotesize
  $^{\dagger}$Corresponding authors\quad Project page: \href{https://super-generative-prior.github.io/papers/hypi2vr}{\texttt{super-generative-prior.github.io/papers/hypi2vr}}
}

\begin{document}

\maketitle

\begin{figure}[h]
    \vspace{-8mm}
    \centering
    \includegraphics[width=\linewidth]{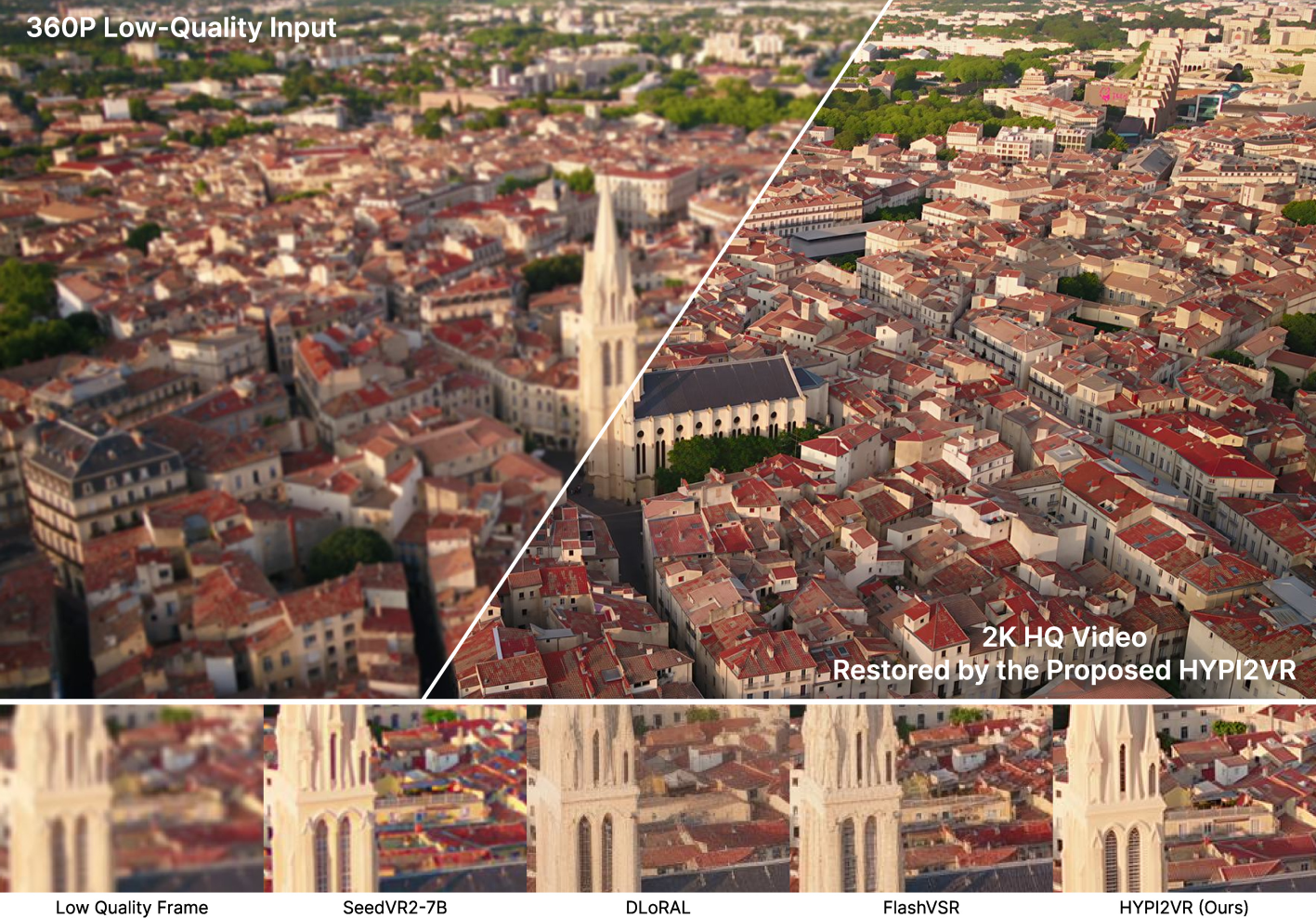}
    \caption{We propose HYPI2VR, which transfers powerful image generation priors to video restoration. Motivated by the stronger generative capacity of image models over video models, HYPI2VR synthesizes fine-grained details while preserving temporal consistency.}
    \label{fig:teaser}
\end{figure}

\begin{abstract}
Recent image generation models produce photo-realistic results with rich details, but transferring their powerful priors to video restoration remains difficult due to temporal inconsistency from stochastic detail synthesis.
Existing video restoration methods improve temporal coherence but often sacrifice visual fidelity, leaving a gap between image-level quality and video-level stability.
We propose a generative video restoration method that bridges this gap by enabling temporally consistent detail synthesis from image generation priors.
Our method combines noise energy rebalancing to suppress structure-disturbing low-frequency stochasticity, motion-aligned noise warping to propagate details along object motion, a self-supervised denoising encoder to mitigate encoder-induced temporal uncertainty, and decoupled temporal attention to model cross-frame dependencies.
Together, these components preserve fidelity, enhance fine-grained realism, and maintain temporal consistency.
Extensive experiments show that our method substantially outperforms existing video restoration approaches in visual quality, detail richness, and temporal stability.
\end{abstract}

\vspace{-3mm}\section{Introduction}\vspace{-1mm}
Image generation and restoration have advanced rapidly, driven by large-scale diffusion models trained on massive high-resolution image datasets~\cite{sdxl, imagen3, FLUX}.
These models now synthesize highly realistic images with fine-grained details.
Video generation has also progressed~\cite{sora, cogvideox, wan, moviegen}, but the scarcity of high-quality video data and the greater complexity of temporal modeling still limit video generative priors.
As a result, current video models continue to trail image models in perceptual quality, especially in resolution, details, and textures.

Advanced photo-realistic image restoration methods \cite{diffbir, supir, faithdiff} commonly build upon pretrained generative models.
However, their stochastic generation makes direct application to video restoration prone to severe temporal inconsistency.
Conversely, existing video restoration methods enforce temporal coherence but often sacrifice visual detail and realism, leaving a substantial quality gap from image generative models.
At the core lies a fundamental tension: high-fidelity restoration requires generative freedom, whereas temporal consistency requires cross-frame constraints.
Our goal is to bridge this gap by transferring powerful image generative priors to video restoration \emph{without compromising either visual quality or temporal consistency}.

Delving into this issue, we identify two fundamental challenges.
First, generative restoration models often suffer from limited fidelity to the low-quality (LQ) input.
Due to their stochastic nature, these models may distort input structures and introduce random deviations.
Second, realistic restoration requires synthesizing missing fine-grained details and textures.
The stochastic generation of such details, when unconstrained across frames, leads to inconsistent hallucinations and manifests as severe temporal artifacts in videos.

To transfer powerful image generative priors to video restoration without sacrificing visual quality, we propose a set of complementary designs that explicitly mitigate the temporal inconsistency introduced by image priors.
Our key observation is that stochastic noise injection controls both detail synthesis and temporal instability.
We therefore rebalance the injected noise in the frequency domain, suppressing low-frequency stochasticity that perturbs image structures while preserving high-frequency noise for realistic detail generation.
To make synthesized details temporally coherent, we further propagate the rebalanced noise along optical flow, so that stochastic detail patterns move consistently with the underlying content.
Beyond injected noise, we find that the encoder itself introduces frame-dependent latent perturbations, which can also cause flickering.
We address this with Noise-to-Noise encoder denoising, using flow-aligned neighboring frames as noisy observations of the same intrinsic latent to suppress encoder-induced uncertainty without clean latent supervision.
Finally, we introduce temporal attention blocks to model cross-frame dependencies beyond flow-based propagation and further stabilize detail synthesis.

Built upon the FLUX prior and the HYPIR framework, HYPI2VR transfers strong image-level generation priors to video restoration while reducing noise-induced, encoder-induced, and cross-frame inconsistency.
With up to 14B parameters, HYPI2VR achieves photo-realistic restoration with stronger fidelity, richer details, and improved temporal consistency over existing video restoration methods, as shown in \figurename~\ref{fig:teaser} and confirmed by extensive evaluations and user studies.

\vspace{-2mm}\section{Motivation and Preliminary}\vspace{-1mm}

\begin{figure}[t]
    \centering
    
    \begin{minipage}[t]{0.34\linewidth}
        \centering
        \includegraphics[width=\linewidth]{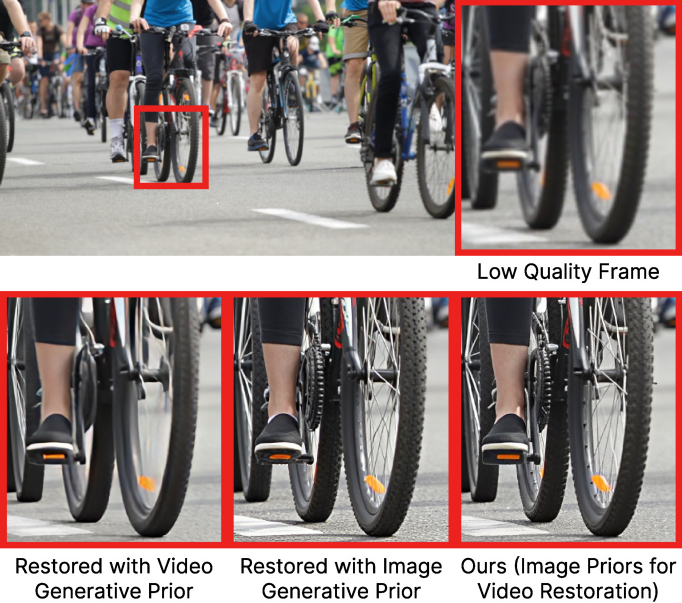}
        \vspace{-3mm}
        \captionof{figure}{Image generation priors enhance video restoration by producing richer details than video-based models, while our approach further improves temporal consistency. Zoom in for a better view.}
        \label{fig:video_prior_vs_image_prior}
    \end{minipage}
    \hfill
    \begin{minipage}[t]{0.29\linewidth}
        \centering
        \includegraphics[width=\linewidth]{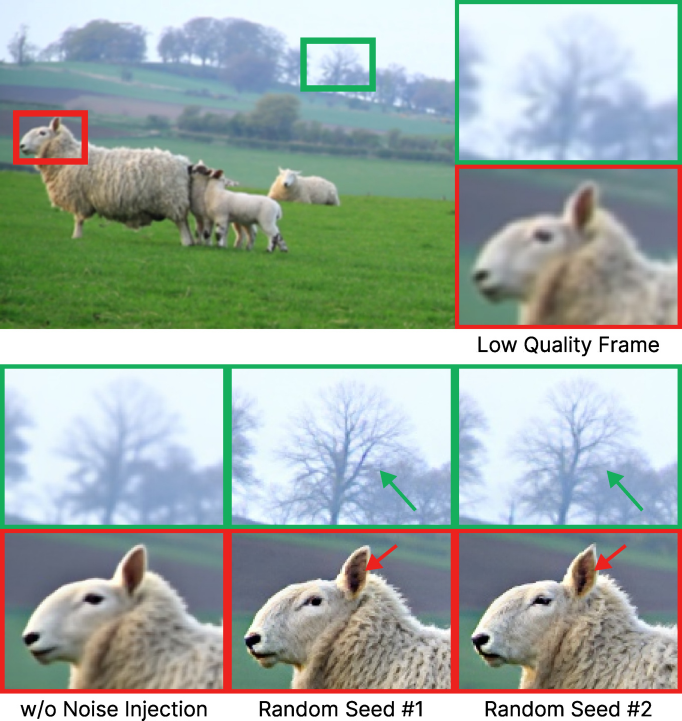}
        \vspace{-3mm}
        \captionof{figure}{Noise injection enhances detail but induces temporal inconsistency, while removing noise improves stability at the cost of detail. Best viewed zoomed in.}
        \label{fig:noise-inject}
    \end{minipage}
    \hfill
    \begin{minipage}[t]{0.34\linewidth}
        \centering
        \includegraphics[width=\linewidth]{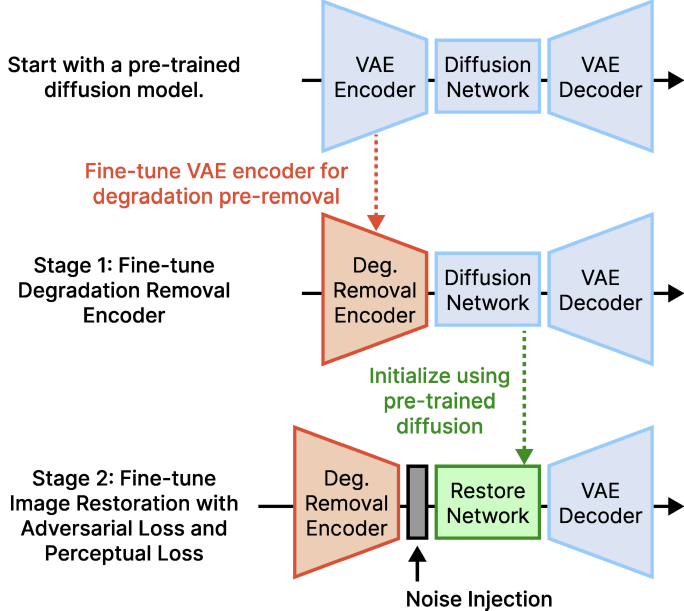}
        \vspace{-3mm}
        \captionof{figure}{Our method builds upon recent efforts \cite{hypir} that perform two-stage adaptation of pre-trained generative models for image restoration, forming the basis for our framework.}
        \label{fig:hypir}
    \end{minipage}
    
    \vspace{-4mm}
\end{figure}

\noindent\textbf{Image Prior \textit{vs.} Video Prior.}\quad
Recent restoration methods increasingly rely on pretrained generative priors~\cite{stablesr,pasd,tsd-sr,faithdiff,supir,diffbir,sr3}.
For video restoration, a natural choice is to use video generative priors~\cite{dam-vsr,flashvsr}, following the intuition that video tasks require video models.
However, we argue that this choice is not necessarily optimal for photo-realistic restoration.
Although video generation has progressed rapidly, current video models still lag behind image models in frame-level visual quality, partly due to the scarcity of high-quality video data and partly because model capacity must be shared between spatial detail synthesis and temporal dynamics modeling.
As a result, video-prior-based restoration often achieves better temporal coherence but weaker spatial richness.
As shown in Fig.~\ref{fig:video_prior_vs_image_prior}, even a strong video model such as WAN~\cite{wan} produces less visually rich restoration results when used as a prior, whereas image-prior-based restoration with FLUX~\cite{FLUX} yields stronger textures and details.
This motivates us to transfer powerful image generative priors to video restoration.

\noindent\textbf{Temporal Inconsistency Issue.}\quad
Although image priors offer stronger visual quality than video priors, applying them to video restoration often causes temporal inconsistency and flickering artifacts.
The key reason is that generative restoration introduces stochastic, hallucinated details: even nearly identical adjacent frames may be restored with noticeably different textures or structures, as shown in Fig.~\ref{fig:noise-inject}.
By contrast, conventional video restoration methods synthesize fewer new details and thus tend to be more temporally stable, revealing a trade-off between detail generation and temporal consistency.
This trade-off can be addressed from two directions: improving video generative priors to match image-level visual fidelity, or exploiting stronger image generative priors while suppressing their temporal inconsistency.
This work follows the latter direction: \emph{Our objective is to transfer powerful image generation priors to video restoration, enabling substantial improvements in visual quality, while mitigating temporal inconsistency.}


\vspace{-1mm}\subsection{Preliminary: Image Restoration with Diffusion Priors}\vspace{-1mm}
\label{sec:preliminaries}
Our method builds on HYPIR~\cite{hypir}, which adapts image generative priors~\cite{FLUX} for efficient image restoration.
We briefly review its pipeline, as illustrated in Fig.~\ref{fig:hypir}.
Given paired data $(I_{\mathrm{deg}}, I)$, where $I_{\mathrm{deg}}$ and $I$ denote the degraded and clean images, the method performs restoration in the latent space of a pretrained FLUX encoder.
It reuses the pretrained VAE decoder $\mathcal{V}_D$, but replaces the original encoder $\mathcal{V}_E$ with a degradation pre-removal encoder $\mathcal{V}_{E_R}$.
This is because $\mathcal{V}_E$ is not robust to severe degradations, so directly encoding $I_{\mathrm{deg}}$ yields latents poorly aligned with clean-image latents.
To obtain $\mathcal{V}_{E_R}$, the encoder is fine-tuned while $\mathcal{V}_D$ is fixed, with the objective
$
\mathcal{L}_{E}=\left\|\mathcal{V}_D(\mathcal{V}_{E_R}(I_{\mathrm{deg}}))-\mathcal{V}_D(\mathcal{V}_{E_R}(I))\right\|_2^2$.
This stage maps the degraded input to a degradation-suppressed latent $\mathbf{y}=\mathcal{V}_{E_R}(I_{\mathrm{deg}})$, which serves as the condition for restoration.

In the second stage, we learn a latent restoration network $\hat{\mathbf{x}}=\mathcal{U}_{\theta}(\mathbf{y})$ and decode the final output as $\hat{I}=\mathcal{V}_D(\hat{\mathbf{x}})$.
Instead of training $\mathcal{U}_{\theta}$ from scratch, we initialize it using a pre-trained diffusion network, i.e., $\theta \leftarrow \theta_{\mathrm{Diff}}$, and perform adversarial fine-tuning for photo-realistic restoration.
Specifically, we train a discriminator $D_{\phi}$ to distinguish restored images from clean images while optimizing $\mathcal{U}_{\theta}$ with an adversarial objective augmented by paired reconstruction terms:
\begin{equation}
\min_{\theta}\max_{\phi}\quad
\mathcal{L}_{\mathrm{adv}}(\theta,\phi)
\;+\;
\lambda_{\mathrm{mse}}\,
\|\mathcal{U}_{\theta}(\mathbf{y})-\mathbf{x}\|_2^2 \;+\;
\lambda_{\mathrm{lpips}}\,
\mathrm{LPIPS}\!\left(\mathcal{V}_D(\mathcal{U}_{\theta}(\mathbf{y})),\, I\right),
\label{eq:prelim_gan}
\end{equation}
where $\mathbf{x}=\mathcal{V}_{E_R}(I)$ denotes the latent of the clean target.
During this training, we keep $\mathcal{V}_{E_R}$ and $\mathcal{V}_D$ fixed and update only $\mathcal{U}_{\theta}$.
In practice, we adopt LoRA \cite{lora} to reduce the number of trainable parameters and improve training efficiency.

In this process, a crucial operation is the injection of stochastic noise into the encoded latent representation, yielding
$
\hat{\mathbf{x}} = \mathcal{U}_{\theta}(\mathbf{y} + \boldsymbol{\epsilon}),
$
where $\boldsymbol{\epsilon}$ denotes the injected noise.
This noise injection transforms the restoration network into a conditional generative process in which the output is conditioned on the input image while retaining stochastic degrees of freedom for detail synthesis.
The injected noise $\boldsymbol{\epsilon}$ directly controls the generation of fine-grained details.
Removing noise injection results in a deterministic restoration process and limits the model's ability to generate high-frequency details.
In contrast, different noise realizations can lead to noticeably different local textures and structures in the restored output, as shown in \figurename~\ref{fig:noise-inject}.
We refer to this operation as \emph{noise injection}.

\begin{figure*}
    \centering
    \includegraphics[width=\linewidth]{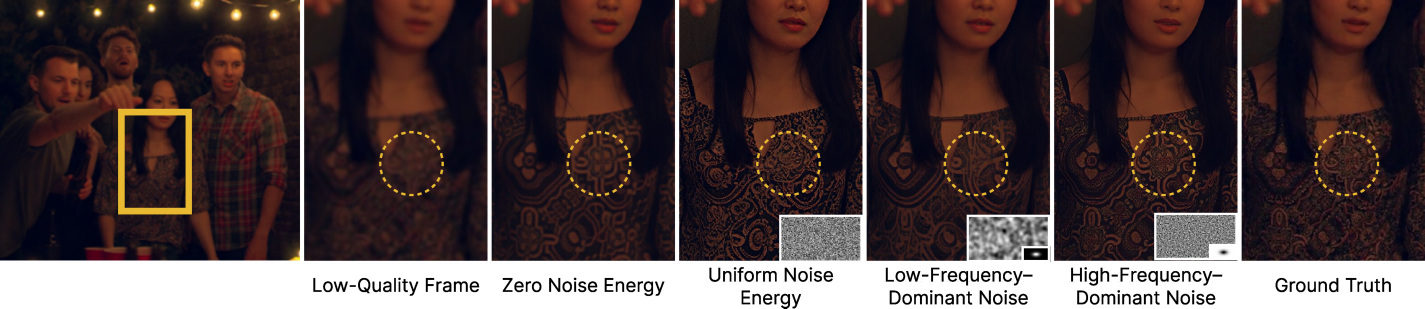}
    \vspace{-5mm}
    \caption{Effect of noise energy distribution on generative image restoration. Allocating noise energy to low frequencies degrades fidelity, while concentrating it in high frequencies preserves structure and enables detailed synthesis.}
    \vspace{-6mm}
    \label{fig:high-pass-noise}
\end{figure*}

\vspace{-2mm}\section{Method}\vspace{-1mm}
\label{sec:mitigating}
%
%
To mitigate temporal inconsistency, it is essential to first understand how it manifests in practice.
We observe that the temporal inconsistency introduced by image generative priors is typically a mixture of two distinct modes:
\begin{itemize}[leftmargin=*, labelsep=0.4em, itemsep=0pt, topsep=0pt, partopsep=0pt, parsep=0pt]
    \item \emph{Fidelity drift with respect to the LQ input frames.} Restoration models may not strictly adhere to the content of the LQ input; instead, they introduce stochastic variations that deviate from the original structure. Such deviations can differ across frames, leading to temporal inconsistency. We refer to this mode as the fidelity issue.
    \item \emph{Inconsistent hallucination of missing details.} LQ frames often lack fine-grained details, forcing the generative model to hallucinate these details during restoration. When such hallucinated details vary across frames, temporal artifacts inevitably emerge.
\end{itemize}

Based on this analysis, our mitigation strategy consists of three components.
First, we improve the fidelity of the restoration, ensuring that the restored results remain faithful to the LQ input frames.
Second, we explicitly regulate the synthesis of additional details by controlling the stochastic components of the generative process.
Third, we promote temporal coherence in the generated details, allowing detail synthesis at the current frame to remain aware of and consistent with neighboring frames.
Together, these three components form the overall method proposed in this work.
In the following subsections, we introduce each component in detail.

\paragraph{Noise Energy Rebalancing for Improved Fidelity.}
Our key observation is that fidelity is governed by the low-frequency content of the LQ inputs.
In practice, LQ inputs already provide reliable low-frequency structures, while missing or corrupted information lies in the high-frequency components.
Therefore, modifying low-frequency content may introduce deviations from the input, leading to fidelity drift.
Instead, restoration should focus on synthesizing high-frequency details while \textbf{preserving} the low-frequency structure of the input.
This observation is closely related to the role of noise injection in generative restoration.
As shown earlier, when noise injection is removed, the restoration model produces largely deterministic outputs that mainly capture low-frequency content.

Motivated by this, we hypothesize that different frequency components of the injected noise $\boldsymbol{\epsilon}$ influence the generation of content at corresponding frequency bands in the restored image.
In particular, high-frequency noise components control the synthesis of fine details, while low-frequency noise components may interfere with the preservation of low-frequency structures from the input.
Based on this insight, we design a noise color rebalancing strategy that suppresses low-frequency noise while retaining high-frequency noise for detail generation.

To control the noise injection on different frequency bands, we analyze the noise in the Fourier domain and decouple it into low-frequency and high-frequency components.
Let $\mathcal{F}(\cdot)$ and $\mathcal{F}^{-1}(\cdot)$ denote the discrete Fourier transform and its inverse, applied channel-wise to latent tensors.
Given an injected noise sample $\boldsymbol{\epsilon} \sim \mathcal{N}(\mathbf{0}, \sigma^2 I)$, we denote its Fourier representation as $\hat{\boldsymbol{\epsilon}}(\mathbf{k}) = \mathcal{F}(\boldsymbol{\epsilon})(\mathbf{k})$,
where $\mathbf{k}=(k_x,k_y)$ indexes the spatial frequency coordinates.
We then introduce a parametric low-pass mask $M_{\mathrm{lp}}(\mathbf{k};\tau)$ as $M_{\mathrm{lp}}(\mathbf{k};\tau)=\exp\!~(-\|\mathbf{k}\|_2^2/\tau^2)$,
where $\tau>0$ controls the characteristic frequency scale of the mask.
The corresponding high-pass mask is defined as its complement $M_{\mathrm{hp}}(\mathbf{k};\tau)=1 - M_{\mathrm{lp}}(\mathbf{k};\tau)=1 - \exp\!~(-\|\mathbf{k}\|_2^2/\tau^2))$.

Using these masks, we decompose the injected noise into low-pass and high-pass components in the Fourier domain:
\begin{equation}
\hat{\boldsymbol{\epsilon}}_{\mathrm{lp}}(\mathbf{k})
=
M_{\mathrm{lp}}(\mathbf{k};\tau)\odot \hat{\boldsymbol{\epsilon}}(\mathbf{k}),
\qquad
\hat{\boldsymbol{\epsilon}}_{\mathrm{hp}}(\mathbf{k})
=
M_{\mathrm{hp}}(\mathbf{k};\tau)\odot \hat{\boldsymbol{\epsilon}}(\mathbf{k}),
\end{equation}
and obtain their spatial-domain counterparts via the inverse Fourier transform as
$\boldsymbol{\epsilon}_{\mathrm{lp}} = \mathcal{F}^{-1}(\hat{\boldsymbol{\epsilon}}_{\mathrm{lp}})$
and
$\boldsymbol{\epsilon}_{\mathrm{hp}} = \mathcal{F}^{-1}(\hat{\boldsymbol{\epsilon}}_{\mathrm{hp}})$.
By construction, the original noise satisfies
$\boldsymbol{\epsilon} = \boldsymbol{\epsilon}_{\mathrm{lp}} + \boldsymbol{\epsilon}_{\mathrm{hp}}$.

Fig.~\ref{fig:high-pass-noise} studies how the frequency distribution of injected noise affects restoration.
Without noise injection, restoration becomes deterministic and closely follows the LQ input, but lacks the stochastic freedom needed for fine-grained detail synthesis.
Uniform noise injection produces richer details, yet also injects substantial low-frequency energy, perturbing global structures and degrading fidelity.
Injecting only the low-frequency component $\boldsymbol{\epsilon}_{\mathrm{lp}}$ further confirms that low-frequency stochasticity contributes little to detail enhancement while disturbing large-scale content.
In contrast, high-frequency noise $\boldsymbol{\epsilon}_{\mathrm{hp}}$ preserves the input's low-frequency structure while enabling realistic detail synthesis, achieving a better fidelity-detail trade-off.

The above decomposition enables the explicit reallocation of noise energy across frequency bands.
We rebalance the injected noise by suppressing its low-frequency components while retaining stochasticity mainly in the high-frequency bands.
We define the rebalanced noise using a frequency-selective high-pass mask as
\begin{equation}
\boldsymbol{\epsilon}^{(\tau)}
=
\mathcal{F}^{-1}\!\Bigl(
M_{\mathrm{hp}}(\mathbf{k};\tau)\odot \hat{\boldsymbol{\epsilon}}(\mathbf{k})
\Bigr).
\label{eq:noise_rebalance_tau}
\end{equation}
The parameter $\tau$ provides continuous control over the amount of low-frequency stochasticity injected into the restoration process.
Smaller values of $\tau$ preserve more low-frequency noise and approach standard noise injection, while larger values of $\tau$ progressively suppress low-frequency perturbations and confine stochasticity to high-frequency details.
In the extreme case $\tau \rightarrow \infty$, stochastic perturbations are fully suppressed.
Finally, the rebalanced noise is injected into the restoration network as
$\hat{\mathbf{x}}
=
\mathcal{U}_{\theta}\!(\mathbf{y} + \boldsymbol{\epsilon}^{(\tau)})$.

\begin{figure}[t]
    \centering
    \includegraphics[width=\linewidth]{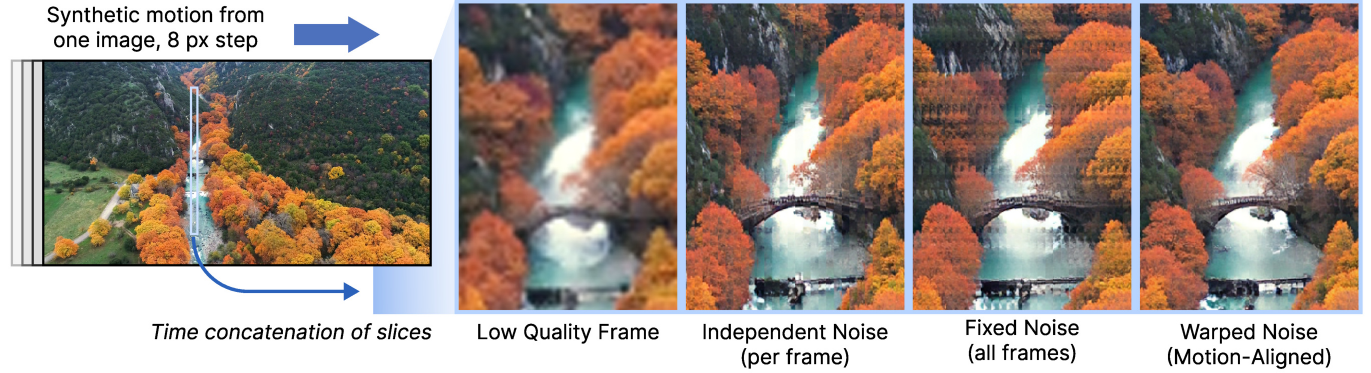}
    \vspace{-2mm}
    \captionof{figure}{Warping noise along motion (8-pixel translation) preserves temporal correspondence, whereas independent or fixed noise causes flickering or repetitive artifacts.}
    \label{fig:noise-fix}
    \vspace{-5mm}
\end{figure}

\paragraph{Noise Warping for Detail Consistency.}
We next address the consistency of synthesized fine-grained details across frames.
High-frequency noise enables realistic detail generation, but unconstrained stochasticity causes flickering and discontinuous textures.
Our key observation is that noise should move with image content: when a region moves to a new location, its synthesized details should remain correlated rather than be re-sampled independently.

We validate this with a controlled synthetic-motion experiment.
Starting from a single image, we generate a sequence by applying a fixed eight-pixel translation, and compare three strategies:
(i) independent noise for each frame,
(ii) fixed noise shared across frames, and
(iii) motion-aligned noise translated together with the image.
As shown in Fig.~\ref{fig:noise-fix}, independent noise causes discontinuous details, fixed noise creates static patterns, while motion-aligned noise preserves stochastic correspondence and yields temporally coherent details.

Based on this insight, we adopt motion-aware noise warping for video restoration, following related ideas in motion-aware video generation~\cite{hiwyn,go-with-the-flow}.
Given degraded frames $I_{\mathrm{deg}}^{t}$ and $I_{\mathrm{deg}}^{t+1}$, we estimate the forward flow $\mathbf{f}_{t \rightarrow t+1}$ and warp the rebalanced noise $\boldsymbol{\epsilon}_t$ from frame $t$ to frame $t{+}1$, so that noise follows the motion of the underlying content.
For robustness, we use correspondence-based transfer: many-to-one mappings are handled by weighted aggregation followed by variance normalization, while uncovered target regions are filled using backward-flow correspondences $\mathbf{f}_{t+1 \rightarrow t}$ when available, or newly sampled Gaussian noise otherwise.
This bidirectional formulation preserves marginal Gaussianity and improves detail consistency under complex motion.

\begin{figure*}
    \centering
    \includegraphics[width=\linewidth]{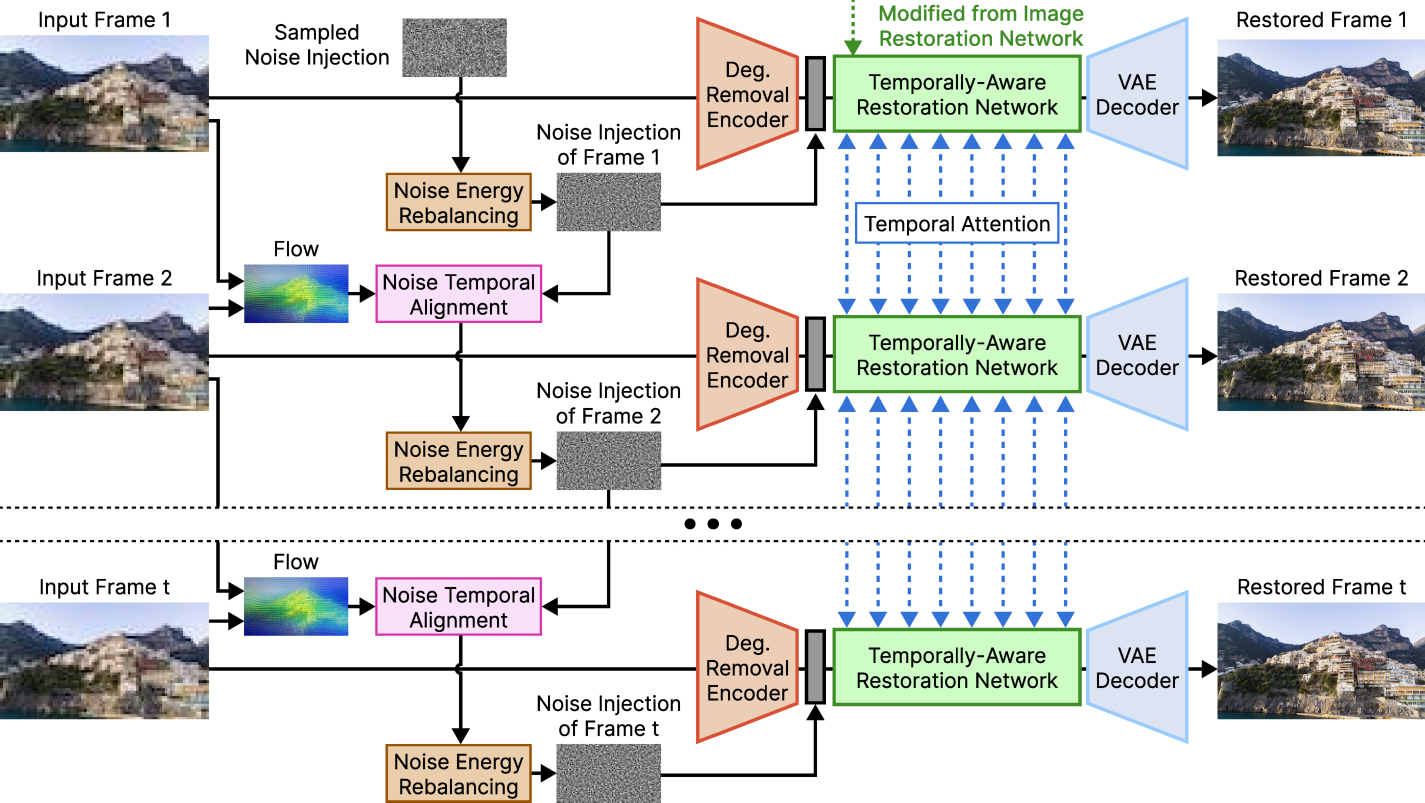}
    \caption{Overview of the proposed video restoration framework.
Noise energy rebalancing, motion-aligned noise propagation, and temporal attention are jointly applied in a sequence-to-sequence manner to achieve temporally consistent restoration.}
    \label{fig:overall}
    \vspace{-4mm}
\end{figure*}

\paragraph{Noise-to-Noise Encoder Denoising.}
Beyond injected noise, we observe that the encoder itself also introduces non-negligible temporal uncertainty.
Following latent image formulations~\cite{kouzelis2025eq}, the encoded latent of a degraded frame can be written as
$
\mathbf{z}^t
=
\mathcal{V}_{E_R}(I_{\mathrm{deg}}^t)
=
\hat{\mathbf{z}}^t+\boldsymbol{\eta}^t+\boldsymbol{\delta}^t
$,
where \(\hat{\mathbf{z}}^t\) denotes the intrinsic content latent, while \(\boldsymbol{\eta}^t\) and \(\boldsymbol{\delta}^t\) denote encoder-induced noise and residual perturbations.
These perturbations vary across adjacent frames and are passed to the restoration network as part of the condition, causing additional flickering.
We therefore train a denoised encoder \(\mathcal{V}_{E_N}\), initialized from \(\mathcal{V}_{E_R}\), while keeping the decoder \(\mathcal{V}_D\) fixed.

Given an input frame \(I_{\mathrm{deg}}^t\), we reconstruct it through the encoder-decoder path:
$
I_{\mathrm{rec}}^t
=
\mathcal{V}_D(\mathcal{V}_{E_N}(I_{\mathrm{deg}}^t))
$.
The pixel reconstruction loss is
$
\mathcal{L}_{\mathrm{mse}}
=
\|I_{\mathrm{rec}}^t-I_{\mathrm{deg}}^t\|_2^2
$.

To suppress frame-dependent encoder perturbations, we further impose a Noise-to-Noise temporal objective in the latent space using optical flow computed from the input frames.
When two aligned observations share the same underlying latent but contain independent zero-mean perturbations, the \(L_2\) optimum is their conditional expectation, i.e., the intrinsic latent rather than the frame-specific noise \cite{lehtinen2018noise2noise}.
Specifically, we first estimate bidirectional flows from \(I_{\mathrm{deg}}\), and use them to warp the reconstructed latent
\(\mathbf{z}_{\mathrm{rec}}^t=\mathcal{V}_{E_N}(I_{\mathrm{deg}}^t)\)
to its neighboring frames, obtaining
\(\mathbf{z}_{\mathrm{rec}}^{t\rightarrow t-1}\) and
\(\mathbf{z}_{\mathrm{rec}}^{t\rightarrow t+1}\).
The neighboring input frames are encoded as
\(\mathbf{z}_{\mathrm{deg}}^{t-1}=\mathcal{V}_{E_N}(I_{\mathrm{deg}}^{t-1})\)
and
\(\mathbf{z}_{\mathrm{deg}}^{t+1}=\mathcal{V}_{E_N}(I_{\mathrm{deg}}^{t+1})\).
The latent flow-based loss is
\begin{equation}
\mathcal{L}_{\mathrm{flow}}
=
\left\|
\mathbf{z}_{\mathrm{rec}}^{t\rightarrow t-1}
-
\mathbf{z}_{\mathrm{deg}}^{t-1}
\right\|_2^2
+
\left\|
\mathbf{z}_{\mathrm{rec}}^{t\rightarrow t+1}
-
\mathbf{z}_{\mathrm{deg}}^{t+1}
\right\|_2^2 .
\end{equation}
The final objective is
$\mathcal{L}_{\mathrm{enc}}
=
\mathcal{L}_{\mathrm{mse}}
+
0.05\,\mathcal{L}_{\mathrm{flow}}$.
This objective preserves current-frame reconstruction in pixel space while enforcing motion-compensated agreement between neighboring latent observations, thereby suppressing frame-dependent encoder perturbations in expectation.
After training, we replace \(\mathcal{V}_{E_R}\) with \(\mathcal{V}_{E_N}\) and use
\(\mathbf{y}^t=\mathcal{V}_{E_N}(I_{\mathrm{deg}}^t)\)
as a more temporally stable condition.

\paragraph{Temporal Attention for Cross-Frame Consistency.}
Noise rebalancing and motion-aware noise warping reduce stochastic inconsistency, but they do not make the restoration backbone temporally equivariant.
In real videos, complex motion, occlusion, and content changes cannot be fully handled by flow-based noise propagation, and the latent restoration network remains sensitive to small input, noise, and alignment variations.
Therefore, we further introduce temporal attention blocks into the DiT backbone to explicitly model cross-frame dependencies.

We adopt a spatial-temporal decoupled attention design.
A temporal attention module~\cite{venhancer,animatediff} is inserted after each spatial attention block, restricting temporal interaction to the time dimension.
This design improves temporal consistency while minimizing interference with spatial attention and single-frame detail synthesis.
Following~\cite{moviegen}, we add sine-cosine positional embeddings along the temporal dimension to provide explicit temporal ordering.
To preserve the image prior's visual quality, we freeze the trained LoRA parameters in the spatial blocks and optimize only the newly introduced temporal blocks.
Since these blocks are trained from scratch, we use a two-stage strategy.
We first train them with per-frame $l_2$ and LPIPS losses, and then add a per-frame adversarial loss initialized from the HYPIR discriminator~\cite{hypir} for generation refinement.
The adversarial objective follows Eq.~\eqref{eq:prelim_gan}, and all training noise is produced by our noise rebalancing and warping process.



\begin{figure*}
    \centering
    \vspace{-2mm}
    \includegraphics[width=\linewidth]{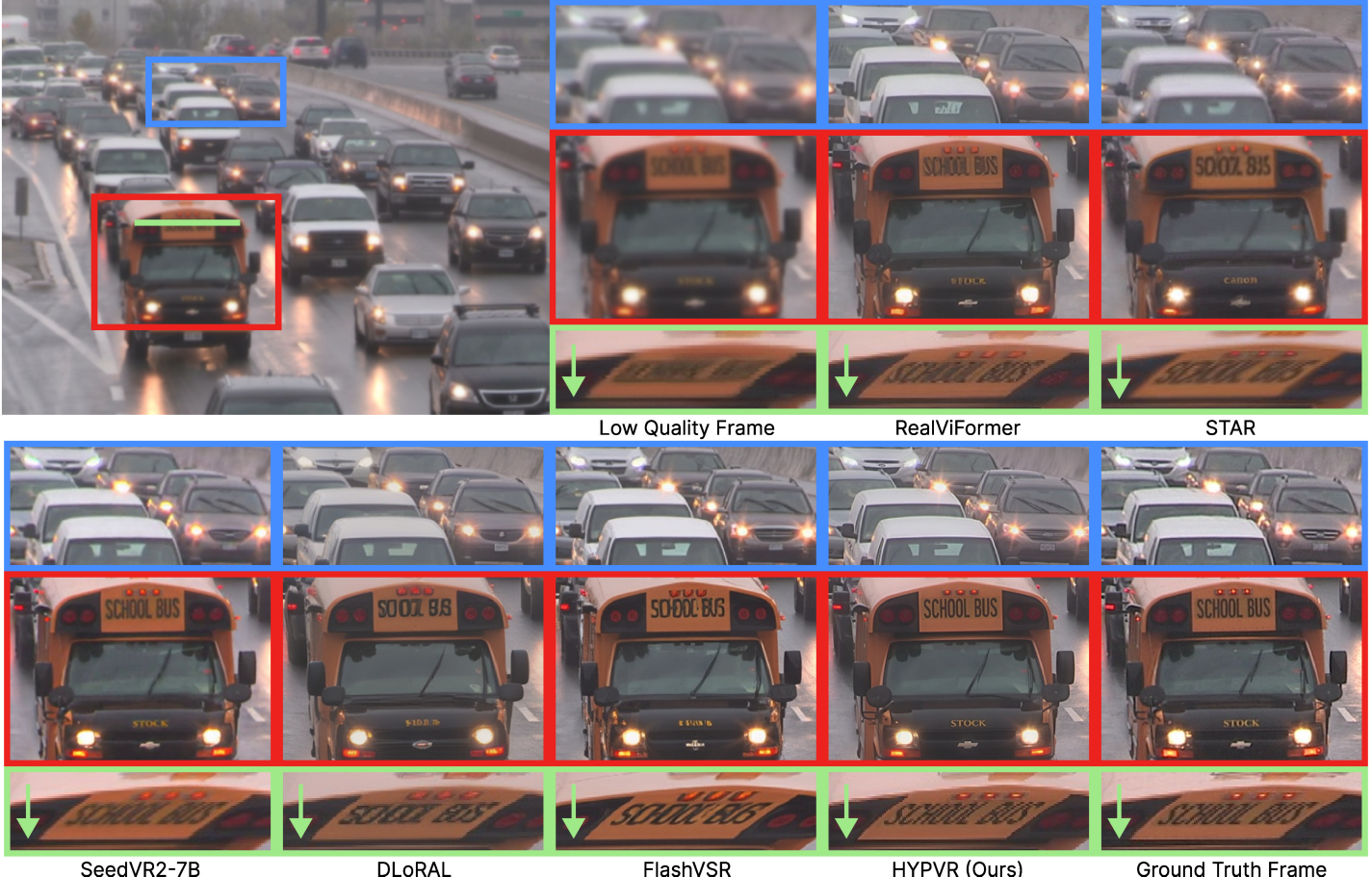}
    \vspace{-4mm}
    \caption{Qualitative and temporal profile comparison. HYPI2VR restores semantically meaningful details while preserving high temporal coherence and visual stability across frames.}
    \label{fig:main_visual_with_temp}
    \vspace{-5mm}
\end{figure*}

\paragraph{Overall Method}
Fig.~\ref{fig:overall} shows our sequence-to-sequence video restoration framework, which takes a 16-frame clip as input and outputs the restored clip during both training and inference.
We use the same high-quality image dataset as HYPIR~\cite{hypir} and $80$K filtered YouTube video clips, each with $64$--$128$ frames at 24~fps, processed at $1024{\times}1024$ resolution.
Unlike SeedVR~\cite{seedvr}, which requires millions of videos, our method uses a much smaller video set.
LQ inputs are synthesized with the RealBasicVSR degradation pipeline~\cite{realbasicvsr}, modified to apply identical blur and sinc kernels across frames and remove extra noise injection, making warped noise the only Gaussian noise source.

Starting from pretrained HYPIR, we first adapt the model to the video-frame distribution for 20{,}000 steps using HYPIR's LoRA training scheme on mixed image-video batches with a $1{:}3$ sampling ratio.
This stage uses RMSProp~\cite{rmsprop} with learning rate $1{\times}10^{-5}$ and batch size 256.
We then freeze the original LoRA parameters and train only the temporal blocks on 16-frame samples.
This stage uses AdamW~\cite{adamw} at $1{\times}10^{-4}$ for 10{,}000 steps, followed by adversarial fine-tuning with RMSProp at $1{\times}10^{-5}$ for 20{,}000 steps.
Training is conducted on four nodes with eight NVIDIA H200 GPUs each, with EMA ratio 0.999, and takes 8 days.

\begin{table*}[t]
\centering
\caption{Quantitative comparisons of different methods on various synthetic and real benchmarks. The best and second performances are marked in \textcolor{best}{red} and \textcolor{second}{blue}, respectively. HYPI2VR consistently achieves competitive or superior results across fidelity, perceptual, and temporal metrics, demonstrating a balanced restoration capability in both synthetic and real-world scenarios.}
\label{tab:main_comp}
\scalebox{0.65}{
\begin{tabular}{c|c|cccccccc}
\toprule
Datasets & Metrics & RealViformer & MGLD-VSR & UAV & STAR & SeedVR2-7B & DLoRAL & FlashVSR & HYPI2VR (Ours)\\
\midrule

\multirow{7}{*}{REDS30}
&PSNR$\uparrow$                 
&\textcolor{best}{24.73} &\textcolor{second}{24.57} &22.56 &22.29 &24.25 &22.77 &21.59 &23.56 \\
&SSIM$\uparrow$                 
&0.6491 &0.6486 &0.5690 &0.5764 &\textcolor{best}{0.6714} &0.5643 &0.5431 &\textcolor{second}{0.6524}\\
&LPIPS$\downarrow$              
&\textcolor{second}{0.2132} &\textcolor{best}{0.2033} &0.5438 &0.5211 &0.2931 &0.2956 &0.3163 &0.2469\\
&MUSIQ$\uparrow$                
&62.30 &63.08 &25.13 &21.80 &51.46 &\textcolor{best}{68.41} &55.68 &\textcolor{second}{64.13} \\
&CLIP-IQA$\uparrow$             
&0.3510 &0.3753 &0.1940 &0.1651 &0.2265 &\textcolor{best}{0.5828} &0.3075 &\textcolor{second}{0.4496}\\
&DOVER$\uparrow$                
&0.3585 &\textcolor{second}{0.4117} &0.1927 &0.1727 &0.3144 &0.3547 &0.3906 &\textcolor{best}{0.4284}\\
&$E_{warp}^*$$\downarrow$
&4.837 &6.066 &\textcolor{best}{3.938} &\textcolor{second}{3.965} &6.527 &7.202 &9.902 &6.727 \\

\midrule

\multirow{7}{*}{UDM10}
&PSNR$\uparrow$                 
&\textcolor{best}{27.89} &\textcolor{second}{27.66} &25.75 &24.47 &27.64 &25.40 &25.46 &26.45 \\
&SSIM$\uparrow$                 
&0.8179 &0.8112 &0.7712 &0.7706 &\textcolor{best}{0.8377} &0.7336 &0.7686 &\textcolor{second}{0.8264}\\
&LPIPS$\downarrow$              
&0.2232 &0.2076 &0.3498 &0.3694 &\textcolor{second}{0.1854} &0.2765 &0.2168 &\textcolor{best}{0.1847}\\
&MUSIQ$\uparrow$                
&61.65 &61.28 &37.12 &29.51 &57.86 &\textcolor{second}{68.04} &61.00 &\textcolor{best}{68.74} \\
&CLIP-IQA$\uparrow$             
&0.4346 &0.4473 &0.2355 &0.1937 &0.4088 &\textcolor{best}{0.6654} &0.4169 &\textcolor{second}{0.5419}\\
&DOVER$\uparrow$                
&0.5668 &0.5667 &0.3740 &0.2885 &0.5557 &0.5613 &\textcolor{second}{0.5703} &\textcolor{best}{0.6477}\\
&$E_{warp}^*$$\downarrow$
&2.514 &3.065 &\textcolor{second}{2.450} &\textcolor{best}{1.836} &2.780 &4.078 &3.471 &2.633\\

\midrule

\multirow{7}{*}{YouHQ40}
&PSNR$\uparrow$                 
&\textcolor{best}{25.64} &\textcolor{second}{25.32} &23.90 &23.45 &25.04 &23.60 &23.51 &24.19 \\
&SSIM$\uparrow$                 
&0.7037 &0.6997 &0.6441 &0.6573 &\textcolor{best}{0.7376} &0.6287 &0.6403 &\textcolor{second}{0.7101}\\
&LPIPS$\downarrow$              
&0.2644 &0.2870 &0.4241 &0.4280 &\textcolor{second}{0.2493} &0.3282 &0.2652 &\textcolor{best}{0.2334}\\
&MUSIQ$\uparrow$                
&64.88 &62.27 &36.10 &29.31 &59.39 &\textcolor{best}{70.14} &62.94 &\textcolor{second}{68.96} \\
&CLIP-IQA$\uparrow$             
&0.4666 &0.4376 &0.2882 &0.2609 &0.4323 &\textcolor{best}{0.6712} &0.4616 &\textcolor{second}{0.5625}\\
&DOVER$\uparrow$                
&0.6968 &\textcolor{second}{0.7085} &0.5047 &0.5305 &0.6748 &0.6670 &0.6854 &\textcolor{best}{0.7418}\\
&$E_{warp}^*$$\downarrow$
&3.364 &3.383 &\textcolor{second}{2.994} &\textcolor{best}{2.135} &3.974 &5.281 &5.029 &4.052 \\

\midrule

\multirow{7}{*}{POUND25}
&PSNR$\uparrow$                 
&25.70 &\textcolor{second}{27.66} &25.70 &25.45 &\textcolor{best}{28.01} &25.32 &24.56 &26.33 \\
&SSIM$\uparrow$                 
&0.7427 &\textcolor{second}{0.7801} &0.7427 &0.7511 &\textcolor{best}{0.8329} &0.7088 &0.7054 &0.7854\\
&LPIPS$\downarrow$              
&0.4081 &0.2625 &0.4081 &0.4016 &\textcolor{best}{0.2232} &0.3679 &0.2732 &\textcolor{second}{0.2377}\\
&MUSIQ$\uparrow$                
&29.34 &56.38 &29.34 &22.12 &43.96 &\textcolor{best}{65.55} &57.11 &\textcolor{second}{61.01}\\
&CLIP-IQA$\uparrow$             
&0.2521 &0.3925 &0.2521 &0.2240 &0.2852 &\textcolor{best}{0.6229} &0.4031 &\textcolor{second}{0.4938}\\
&DOVER$\uparrow$                
&0.4291 &0.6155 &0.6002 &0.4020 &0.5573 &0.5969 &\textcolor{second}{0.6198} &\textcolor{best}{0.6830}\\
&$E_{warp}^*$$\downarrow$
&3.425 &3.669 &\textcolor{second}{2.792} &\textcolor{best}{2.472} &3.435 &5.094 &5.761 &3.812 \\

\midrule
\multirow{3}{*}{VideoLQ}
&MUSIQ$\uparrow$                
&50.43 &48.07 &24.76 &29.52 &35.09 &\textcolor{best}{62.78} &49.86 &\textcolor{second}{56.15} \\
&CLIP-IQA$\uparrow$             
&0.3450 &0.3103 &0.2152 &0.2214 &0.2222 &\textcolor{best}{0.6252} &0.3477 &\textcolor{second}{0.4078}\\
&DOVER$\uparrow$                
&0.4962 &0.5148 &0.3241 &0.4299 &0.4344 &\textcolor{second}{0.5531} &0.5346 &\textcolor{best}{0.5592}\\
\bottomrule
\end{tabular}
}
\vspace{-3mm}
\end{table*}

{
\setlength{\tabcolsep}{6pt}
\begin{table*}[t]
\centering
\vspace{1mm}
\caption{Ablation study of key modules on the POND25 dataset. 3D-SW and 1D denote the 3D shift-window attention block and the 1D full attention block used for temporal modeling. The best and second performances are marked in \textcolor{best}{red} and \textcolor{second}{blue}, respectively. Each component contributes to the overall performance trade-off across fidelity, perceptual quality, and temporal consistency.}
\label{tab:ab_all}
\scalebox{0.65}{
\begin{tabular}{ccccc|ccccccc}
\toprule
Noise & Noise & Temp. & Temp. & Feat.&\multirow{2}{*}{PSNR$\uparrow$}&\multirow{2}{*}{SSIM$\uparrow$}&\multirow{2}{*}{LPIPS$\downarrow$}&\multirow{2}{*}{MUSIQ$\uparrow$}&\multirow{2}{*}{CLIP-IQA$\uparrow$}&\multirow{2}{*}{DOVER$\uparrow$}&\multirow{2}{*}{$E_{warp}^*\downarrow$} \\
Rebal. & Align. & Encode & Block & Prop.&&&&&&& \\
\midrule

 &  &  &  & 
& 23.90 
& 0.6826 
& 0.3421 
& 61.80 
& \textcolor{second}{0.5035}
& \textcolor{second}{0.6896}
& 10.061 \\

 & \checkmark &  &  & 
& 23.89
& 0.6791 
& 0.3474
& 61.69
& \textcolor{best}{0.5069}
& \textcolor{best}{0.6956}
& 7.648 \\

\checkmark &  &  &  & 
& 25.80 
& 0.7428 
& 0.2663 
& \textcolor{best}{62.08}
& 0.4729 
& 0.6590 
& 7.048 \\

\checkmark & \checkmark &  &  & 
& 25.81 
& 0.7428 
& 0.2663 
& \textcolor{second}{62.01}
& 0.4725 
& 0.6653 
& 5.670 \\

\checkmark & \checkmark & \checkmark & & 
& 25.59
& 0.7571 
& 0.2547 
& 61.70
& 0.4603 
& 0.6858 
& 5.340 \\

\checkmark & \checkmark & \checkmark & 3D-SW & 
& 25.04
& 0.7370
& 0.2788
& 49.57
& 0.3283
& 0.6395
& 5.088 \\

\checkmark & \checkmark & \checkmark & 1D & \checkmark
& \textcolor{best}{27.80}
& \textcolor{best}{0.7979}
& \textcolor{best}{0.2369}
& 53.20
& 0.4098
& 0.6560
& \textcolor{best}{3.385} \\

\checkmark & \checkmark & \checkmark & 1D & 
& \textcolor{second}{26.33} 
& \textcolor{second}{0.7854} 
& \textcolor{second}{0.2377}
& 61.01
& 0.4938 
& 0.6830
& \textcolor{second}{3.812} \\

\bottomrule
\end{tabular}
}
\vspace{-4mm}
\end{table*}
}

\vspace{-2mm}\section{Experiments}\vspace{-1mm}
Following SeedVR2~\cite{seedvr2}, we evaluate on UDM10~\cite{udm10}, REDS30~\cite{reds30}, and YouHQ40~\cite{uav}.
We further curate \textit{POND25} to examine dynamic camera motion and complex real-world scenes, selecting 25 high-quality clips from POND\footnote{https://www.pond5.com/} with an average duration of 4 seconds and resolution of $2560{\times}1440$.
All clips include complete copyright information and are permitted for display.
Synthetic LQ inputs use the same degradation process as temporal-module training, and real-world evaluation uses VideoLQ~\cite{realbasicvsr}.
We report PSNR~\cite{psnr}, SSIM~\cite{ssim}, and LPIPS~\cite{lpips} for fidelity; CLIP-IQA~\cite{clipiqa}, MUSIQ~\cite{musiq}, and DOVER~\cite{dover} for perceptual quality; and warping error~\cite{ewarp} for temporal consistency.
Wavelet-based color correction is applied to all restored results before metric computation.
For inference, we perform latent-space 3D tiling with spatial tile size/stride $1024/256$ and temporal tile size/stride $16/4$.
Inputs shorter than 1024 pixels in height are padded before restoration and resized back afterward.
We fix $\tau=1.0$, consistent with training.
Baselines include RealViFormer~\cite{realviformer}, multi-step diffusion methods UAV~\cite{uav}, MGLD-VSR~\cite{mgld}, STAR~\cite{star}, and one-step diffusion-pretrained methods SeedVR2-7B~\cite{seedvr2}, DLoRAL~\cite{dloral}, and FlashVSR~\cite{flashvsr}.

\vspace{-1mm}\subsection{Results}\vspace{-1mm}
We evaluate HYPI2VR on synthetic and real-world video restoration benchmarks.
Due to space limitations, we show representative visual comparisons in Fig.~\ref{fig:main_visual_with_temp}, with more results on synthetic videos, real-world videos, historical films, and motion-blur cases provided in the appendix.
As shown in Fig.~\ref{fig:main_visual_with_temp}, HYPI2VR restores semantically faithful fine details while maintaining stable temporal profiles.
Textual structures such as the ``School Bus'' label are clearly reconstructed across frames without obvious flickering or geometric distortion.
Compared with existing methods, HYPI2VR produces richer textures and sharper structures, demonstrating the benefit of combining powerful image generative priors with our temporal consistency design.
More qualitative comparisons are provided in Sec.~\ref{sec:appendix_more_results}.

Tab.~\ref{tab:main_comp} reports quantitative comparisons.
HYPI2VR achieves competitive performance across fidelity, perceptual quality, and temporal consistency metrics.
For SSIM and LPIPS, it outperforms most diffusion-based methods and approaches reconstruction-oriented models, showing that noise energy rebalancing improves detail synthesis while preserving structural fidelity.
For MUSIQ, CLIP-IQA, and DOVER, HYPI2VR obtains strong and balanced perceptual scores on both synthetic and real-world datasets.
For temporal consistency, HYPI2VR provides a better trade-off between stable motion and high-frequency detail preservation.
Although some methods report lower warping error, they often produce over-smoothed outputs, while HYPI2VR maintains sharper and more realistic details.

Finally, we emphasize that quantitative metrics may not fully reflect the quality of generative restoration.
They can favor excessive high-frequency responses, even when the results are semantically or physically implausible, such as over-sharpened motion-blur regions.
HYPI2VR instead preserves natural blur when appropriate and restores clarity in salient structures, leading to more faithful and coherent videos.
This is further verified by the user study in the appendix, where HYPI2VR achieves the best rank score and the highest MOS, substantially outperforming all competing methods.

\begin{figure*}
    \centering
    \includegraphics[width=\linewidth]{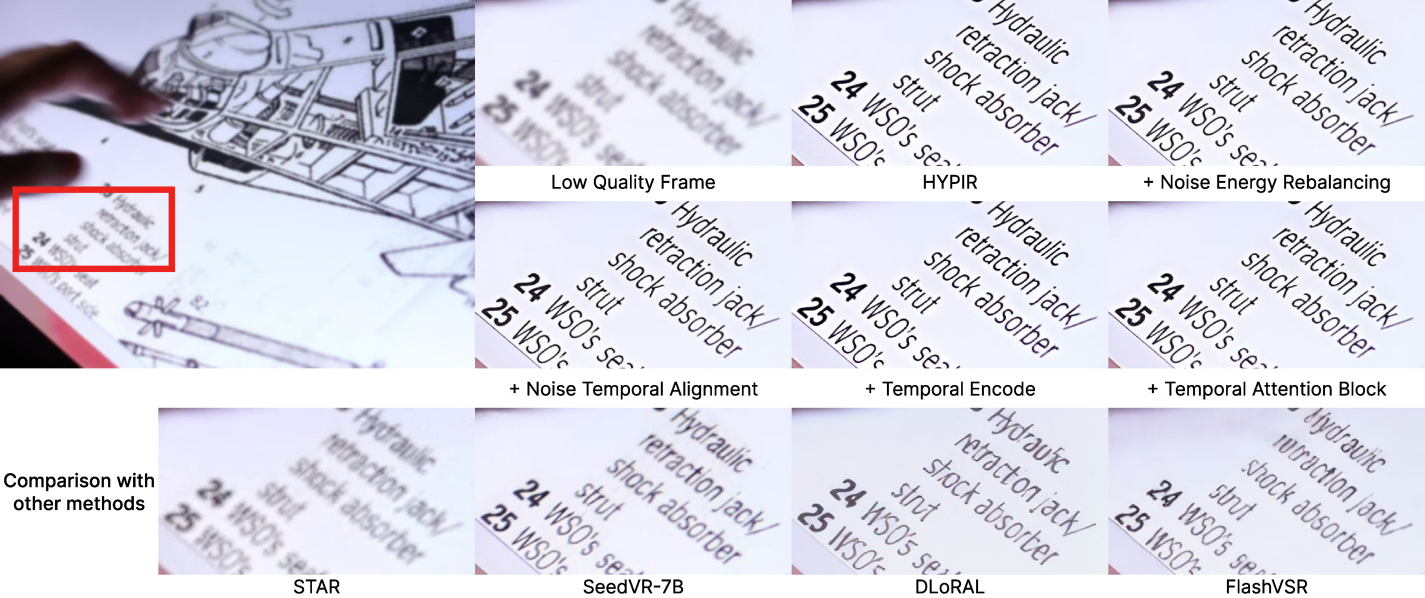}
    \vspace{-5mm}
    \caption{The proposed modules improve temporal consistency while preserving text fidelity and high-quality texture synthesis, showing that better alignment does not compromise perceptual detail or generative capability.}
    \vspace{-4mm}
    \label{fig:ablation_visual}
\end{figure*}

\vspace{-1mm}\subsection{Ablation Study}\vspace{-1mm}
We conduct ablation studies to validate each component of \textbf{HYPI2VR}, with results summarized in Tab.~\ref{tab:ab_all}.
Starting from the HYPIR baseline, which shows limited fidelity and poor temporal stability, Noise Energy Rebalancing substantially improves spatial fidelity, increasing PSNR from 23.90 dB to 25.80 dB and reducing LPIPS to 0.2663.
This confirms that suppressing low-frequency noise better preserves reliable input structures while retaining high-frequency stochasticity for detail synthesis.
Adding Noise Temporal Alignment further improves temporal stability, reducing $E_{warp}^*$ from 7.048 to 5.670 with little loss in image quality, showing the benefit of aligning stochastic details across frames.
Temporal Encode provides additional motion guidance and further improves consistency with minimal impact on fidelity.
Finally, introducing the Temporal Block explicitly models inter-frame dependencies and improves the overall balance among fidelity, perceptual quality, and temporal consistency.
The qualitative results in Fig.~\ref{fig:ablation_visual} further show that our final model achieves strong temporal coherence while preserving fine details.
We also study alternative temporal modeling designs and feature propagation.
As shown in Tab.~\ref{tab:ab_all}, replacing 1D temporal attention with 3D shift-window attention or adding feature propagation leads to inferior visual quality, supporting our final design choice.
Additional ablation results, visualizations, and detailed ablation settings are provided in the supplementary material.

\vspace{-1mm}\subsection{User Study}\vspace{-1mm}

\begin{figure}[t]
    \centering
    \vspace{-4mm}
    
    \begin{minipage}[t]{0.49\linewidth}
        \centering
        \includegraphics[width=\linewidth]{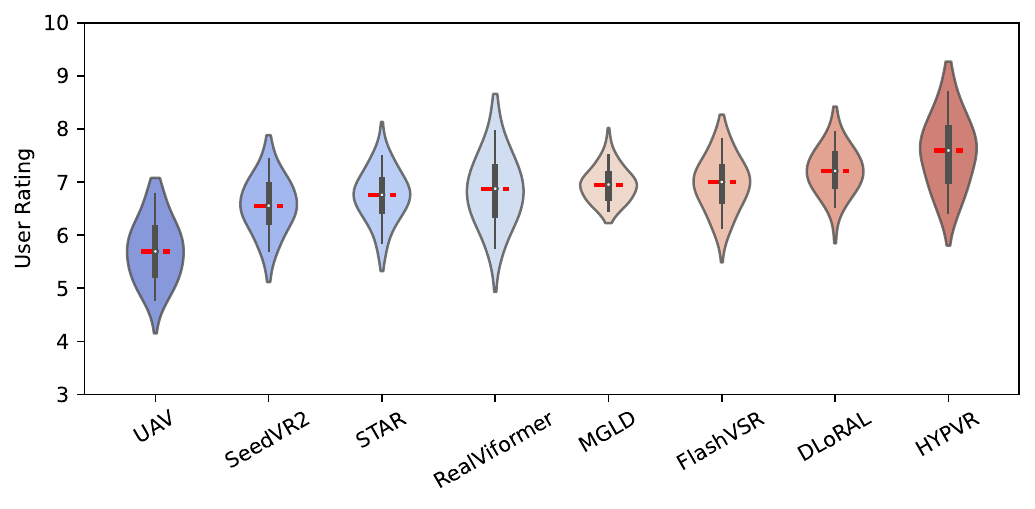}
        \includegraphics[width=\linewidth]{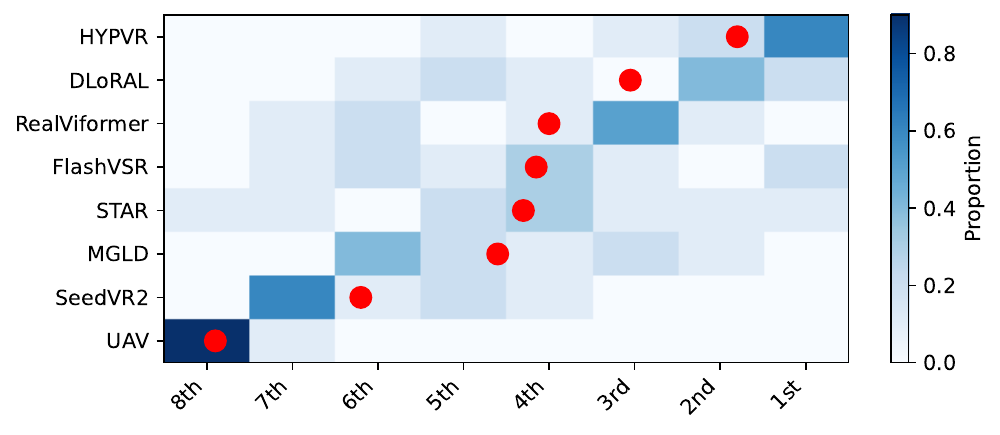}
        \captionof{figure}{User study on \textbf{video-level} evaluations. Top: Violin plots of user ratings (1--10). Bottom: Ranking distribution across methods (1st to 8th). HYPI2VR achieves the highest perceptual scores and dominates top-ranked preferences.}
        \label{fig:user_video}
    \end{minipage}
    \hfill
    \begin{minipage}[t]{0.49\linewidth}
        \centering
        \includegraphics[width=\linewidth]{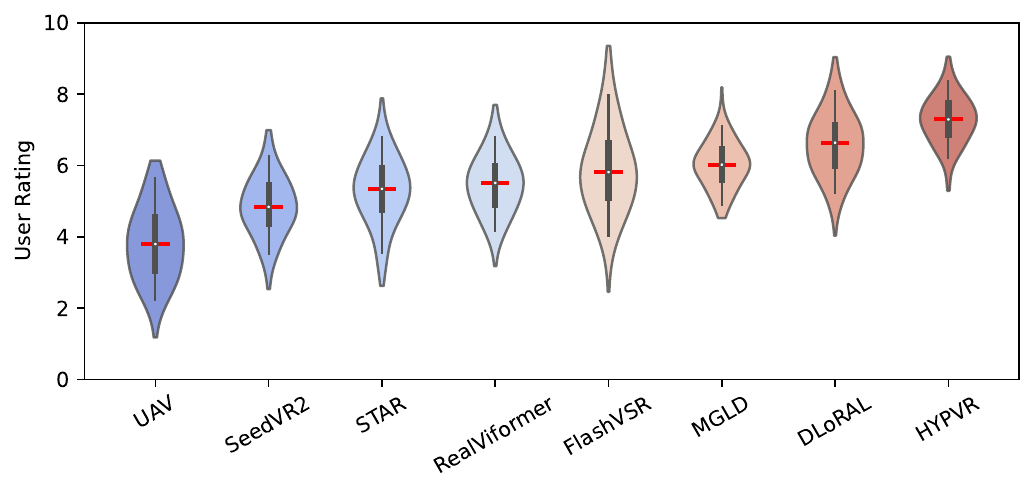}
        \includegraphics[width=\linewidth]{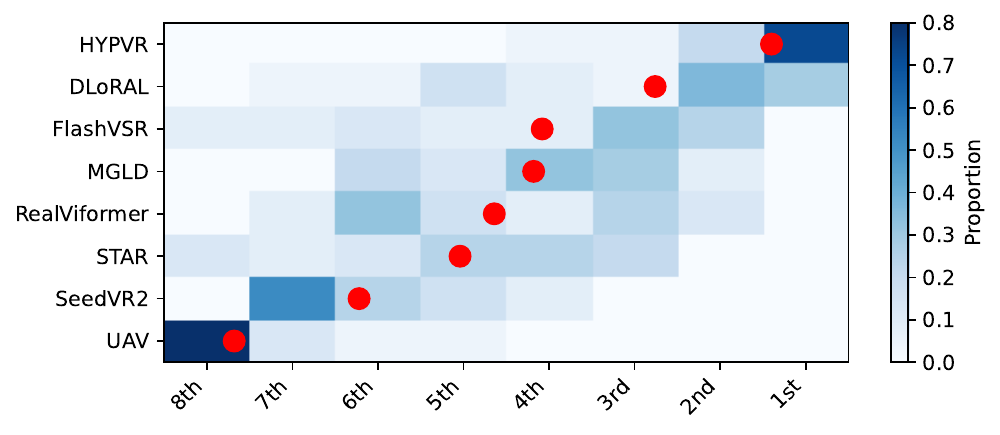}
        \captionof{figure}{User study on \textbf{frame-level} evaluations. Top: Violin plots of user ratings. Bottom: Ranking distribution across methods. HYPI2VR maintains strong visual quality, outperforming both generative and non-generative baselines.}
        \label{fig:user_image}
    \end{minipage}
    
    \vspace{-5mm}
\end{figure}

To evaluate the performance of HYPI2VR, we conduct a user study on ten randomly selected scenes drawn from both historical films and the VideoLQ dataset.
The study consists of two parts.
First, participants evaluate restored video clips at 2K resolution, each with a duration of 4–6 seconds, produced by different methods.
Second, static frames are extracted from the same scenes for frame-level quality assessment.
All participants, none of whom have formal training in computer vision, are asked to rate each video or image on a scale from 1 to 10 based on overall visual quality.

We recruit 100 participants via Prolific and collect a total of 2,480 video ratings and 6,400 frame ratings.
As shown in \figurename~\ref{fig:user_video}, HYPI2VR achieves the highest average perceptual score among all evaluated models.
While SeedVR2 demonstrates strong structural recovery and temporal stability, its perceptual score is lower due to limited fine-detail reconstruction.
In contrast, FlashVSR, DLoRAL, and our method obtain higher perceptual ratings despite exhibiting relatively larger warp errors, suggesting that fine-detail preservation is an important factor in human perception, and excessive emphasis on temporal consistency may degrade perceived quality.

For static frame evaluation, as shown in \figurename~\ref{fig:user_image}, our method consistently attains the highest visual quality scores across all tested scenes.
FlashVSR shows larger variance in user ratings, performing well in low-motion scenarios but degrading noticeably in highly dynamic scenes, whereas our method maintains stable performance across diverse conditions.
Interestingly, the non-generative approach RealViFormer achieves ratings comparable to generative methods.
We attribute this to the absence of a generative prior, which helps avoid semantic distortions and hallucinated artifacts, resulting in cleaner and more faithful restorations.

\vspace{-2mm}\section{Conclusion}\vspace{-1mm}
In this work, we present HYPI2VR, a video restoration framework that tries to bridge the gap between visual quality and temporal consistency.
%
%
Extensive experiments and user studies demonstrate that our approach achieves vivid, detail-rich restoration while maintaining strong temporal stability, offering a promising direction for high-quality generative video restoration.

\paragraph{Acknowledgment.} This work was partially supported by the National Natural Science Foundation of China (Grant No.62276251) and  the Joint Lab of CAS-HK. This research was partially funded by the Ministry of Education and Science of Bulgaria (support for INSAIT, part of the Bulgarian National Roadmap for Research Infrastructure).

\medskip

\bibliographystyle{unsrt}
\bibliography{reference}

\clearpage

\appendix

\section{Related Work}
\noindent\textbf{Video Restoration Networks}
Early research works in video restoration \cite{vespcn,duf,edvr,tdan,rlsp,rbpn,basicvsr,basicvsr++,psrt,vrt,mia-vsr} primarily focus on refining network architectures, loss functions, and degradation synthesis strategies to reconstruct continuous high-quality (HQ) frames from their corresponding LQ inputs.
DUF \cite{duf} applies 3D convolutions to model spatio-temporal relationships.
EDVR and TDAN \cite{edvr,tdan} propose deformable convolution-based feature alignment with motion compensation.
RLSP \cite{rlsp} implicitly captures temporal dependencies through hidden states, while RBPN \cite{rbpn} further incorporates a recurrent projection module for frame prediction.
BasicVSR \cite{basicvsr} and BasicVSR++ \cite{basicvsr++} adopt an optical flow-based bidirectional propagation mechanism to aggregate temporal information.
The emergence of vision Transformer networks \cite{vit} enables the modeling of long-range spatio-temporal dependencies through attention mechanisms.
VRT \cite{vrt} is the first to introduce this framework to video restoration, and subsequent works design efficient patch-level alignment \cite{psrt} or masked intra-frame and inter-frame attention modules \cite{mia-vsr,realviformer} to reduce redundant computations in Transformer-based architectures.
However, these methods largely rely on idealized degradation models, leading to a synthetic-to-real domain gap and resulting in over-smoothed outputs rather than photo-realistic restoration results.

\noindent\textbf{Generative Prior for Restoration}
Generative priors have demonstrated remarkable capabilities in modeling complex natural video distributions and synthesizing rich details and textures, thereby improving the perceptual quality of restored images/videos, particularly in the presence of complex real-world degradations.
The corresponding models can be broadly divided into two paradigms: generative adversarial network (GAN) \cite{gan}-based and diffusion-based methods.
Both paradigms have been extensively explored in the context of image restoration and have led to a number of highly successful models.

GANs are among the earliest effective generative models for natural image synthesis. From a theoretical perspective, adversarial training of GANs can be interpreted as minimizing divergences between generated and real data distributions and has been widely adopted in image restoration tasks \cite{esrgan,real-esrgan,pan2021exploiting,chan2021glean,zhu2022disentangled,ledig2017photo}. When extended to video restoration, the inherent training instability and mode collapse of GAN-based models pose additional challenges. Early GAN-based video restoration studies primarily focus on enhancing visual realism by incorporating adversarial learning into video super-resolution frameworks \cite{vsrresfeatgan}, demonstrating that adversarial supervision can significantly improve perceptual quality.
As awareness of temporal inconsistency issues has grown, subsequent approaches \cite{tecogan,longvideogan} introduce explicit temporal constraints.
%
%
More recent studies explore scaling up GAN models to further enhance detail synthesis.
VideoGigaGAN \cite{videogigagan} adapts large-capacity GAN architectures originally developed for image generation to the video domain by introducing temporal alignment and consistency mechanisms.
%
%
Despite these advances, GAN-based video restoration methods often suffer from suboptimal generative quality and limited robustness when confronted with complex and unknown real-world degradations.

Diffusion-based models \cite{ddpm, ddim, ldm, dhariwal2021diffusion} progressively add stochastic noise via a forward diffusion process and recover clean samples through a reverse denoising process, thereby modeling the underlying data distribution. 
Building on this generative paradigm, diffusion-based approaches have recently advanced rapidly and become a mainstream research direction not only for image generation but also for image restoration \cite{stablesr, sr3, diffbir, supir, pasd, tsd-sr, faithdiff}, demonstrating strong performance in recovering high-quality and faithful images.
Nevertheless, the inherent stochasticity of generative diffusion models makes them difficult to directly apply to frame-wise video processing without introducing temporal inconsistency.
Early attempts employ optical flow-based alignment mechanisms \cite{mgld, stablevsr} or incorporate trainable temporal modules \cite{uav, sateco}.
Subsequently, DiffVSR \cite{diffvsr} uses Interweaved Latent Transition to ensure strong temporal consistency with no additional training cost.
Despite these advances, current latent diffusion model (LDM) \cite{ldm}-based video restoration methods still struggle to jointly achieve strong temporal consistency and photo-realistic visual quality.

Diffusion Transformers (DiT) \cite{dit} represent an advance in text-to-video generation \cite{moviegen}, enabling the synthesis of HQ videos with improved temporal stability.
DiT-based networks capture continuous dynamics in long videos by modeling temporal information directly within a Transformer, thereby ensuring temporal consistency and smooth motion.
In addition, their high-capacity architectures better preserve scene details and complex dynamics.
Recent frameworks leverage such large-scale diffusion priors, including SD3.5, FLUX \cite{FLUX}, CogVideoX \cite{cogvideox}, and WAN \cite{wan}, to address image and video restoration tasks.
For example, HYPIR \cite{hypir} initializes an image restoration model from a pre-trained diffusion prior and then fine-tunes it with adversarial training, achieving strong performance within a single forward pass.
This paradigm has subsequently been extended to video restoration.
SeedVR \cite{seedvr, seedvr2} optimizes the attention mechanism in DiT to improve computational efficiency and further performs adversarial post-training using real-world data.
STAR \cite{star} adds a pre-attention module to improve local details and suppress degradation artifacts.
DLoRAL \cite{dloral} proposes a dual LoRA \cite{lora} learning paradigm to extract temporal priors from LQ inputs.
Through staged distillation training, FlashVSR \cite{flashvsr} can be reliably extended to ultra-high-resolution video restoration.

\section{Additional Details of the Method}
\label{sec:additional_analysis}

\subsection{Analysis of Encoder-Induced Temporal Noise}
\label{sec:encoder_noise}

\begin{figure}[t]
    \centering
    \includegraphics[width=\linewidth]{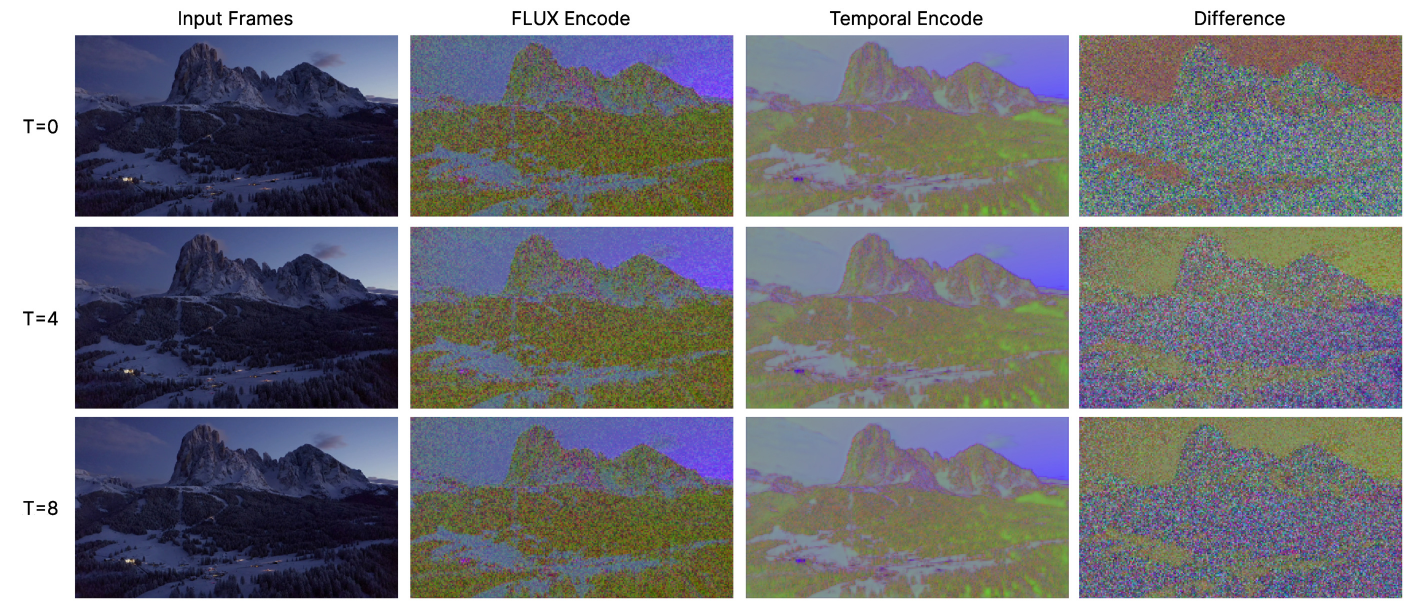}
    \vspace{-3mm}
    \caption{
    Visualization of encoder-induced temporal inconsistency in latent space. 
    We project 16-channel latents to 3 channels using PCA, followed by per-sample min--max normalization for visualization. 
    The original encoder produces temporally inconsistent noise patterns across frames, while our temporal encoder yields more stable and consistent representations. 
    The difference maps further highlight frame-dependent noise variations. 
    }
    \vspace{-3mm}
    \label{fig:temporal_encode_vis}
\end{figure}

To validate our intuition that the encoder introduces frame-dependent perturbations, we conduct a toy experiment to visualize latent representations across consecutive frames.
Given input frames $\{I_{\mathrm{deg}}^t\}$, we extract their latent features using both the original encoder $\mathcal{V}_{E_R}$ and our temporal encoder $\mathcal{V}_{E_N}$.
To make the latent patterns observable, we project the 16-channel latent features to 3 channels using PCA, followed by min-max normalization.

As shown in Fig.~\ref{fig:temporal_encode_vis}, the original encoder produces noticeable variations in latent representations across frames, despite the underlying scene remaining largely unchanged.
These variations correspond to frame-dependent noise patterns introduced by the encoder.
In contrast, the temporal encoder produces significantly more consistent latent representations over time.
The difference maps further reveal that these inconsistencies are primarily localized in stochastic patterns rather than structural content.
These observations provide empirical evidence that encoder-induced perturbations contribute to temporal instability, and demonstrate that our temporal encoding design mitigates such noise.

\subsection{Detailed Experimental Settings for Noise Analysis}
\label{sec:noise_analysis}

\paragraph{Effect of Noise Energy Distribution (Fig.~\ref{fig:high-pass-noise}).}
To isolate the effect of noise energy distribution, all results in Fig.~\ref{fig:high-pass-noise} are obtained from independently trained HYPIR models from scratch.
Each model is trained using a specific noise configuration, while all other training settings are kept identical.
During inference, the same noise distribution as used in training is applied, ensuring no out-of-distribution (OOD) mismatch between training and testing.

\paragraph{Effect of Temporal Noise Consistency (Fig.~\ref{fig:noise-fix}).}
In Fig.~\ref{fig:noise-fix}, we analyze the impact of temporal noise consistency using the official HYPIR-FLUX model, where the only varying factor is the inter-frame relationship of noise.
Specifically, \emph{independent noise} samples noise independently for each frame, \emph{fixed noise} shares an identical noise map across all frames, and \emph{warped noise} aligns the noise with pixel motion.
For the warped setting, we apply a translation of 8 pixels, corresponding to an integer shift in latent space, allowing a clear demonstration of motion-aligned noise propagation.

\section{Dataset Details}

\subsection{Video training Dataset}
To construct a high-quality training dataset, we collect approximately 2{,}000 hours of high-resolution videos from YouTube, covering diverse categories such as portraits, architecture, animals, plants, and natural scenes. The videos range between 4K and 8K resolution and maintain frame rates above 24 FPS. Only sequences with sufficiently high bitrates and no observable compression artifacts are retained.
Before scene segmentation, we remove videos with black borders and discard those with an effective frame height below 2{,}048 pixels. Scene boundaries are detected using the PySceneDetect toolkit. After filtering out scenes shorter than 64 frames at 24 FPS, we obtain a total of $840$K clips.
Each clip is evaluated from two perspectives: (1) \textit{frame quality} and (2) \textit{scene dynamics}. Frame quality is computed by sampling frames at 3 FPS and averaging MUSIQ and LAION-Aesthetic scores. Scene dynamics are quantified using the optical flow magnitude computed by RAFT to represent temporal motion intensity. By jointly ranking clips with respect to both criteria, we obtain a curated subset of $62$K high-quality video clips, with a retention rate of $7.38\%$.
All selected clips are downsampled by a factor of two to enhance frame fidelity. The resulting high-quality frames are extracted at 24 FPS and stored as individual images for subsequent training.

\subsection{Composition of the POND5 Validation Dataset}

\begin{figure*}[hb]
    \centering
    \includegraphics[width=\linewidth]{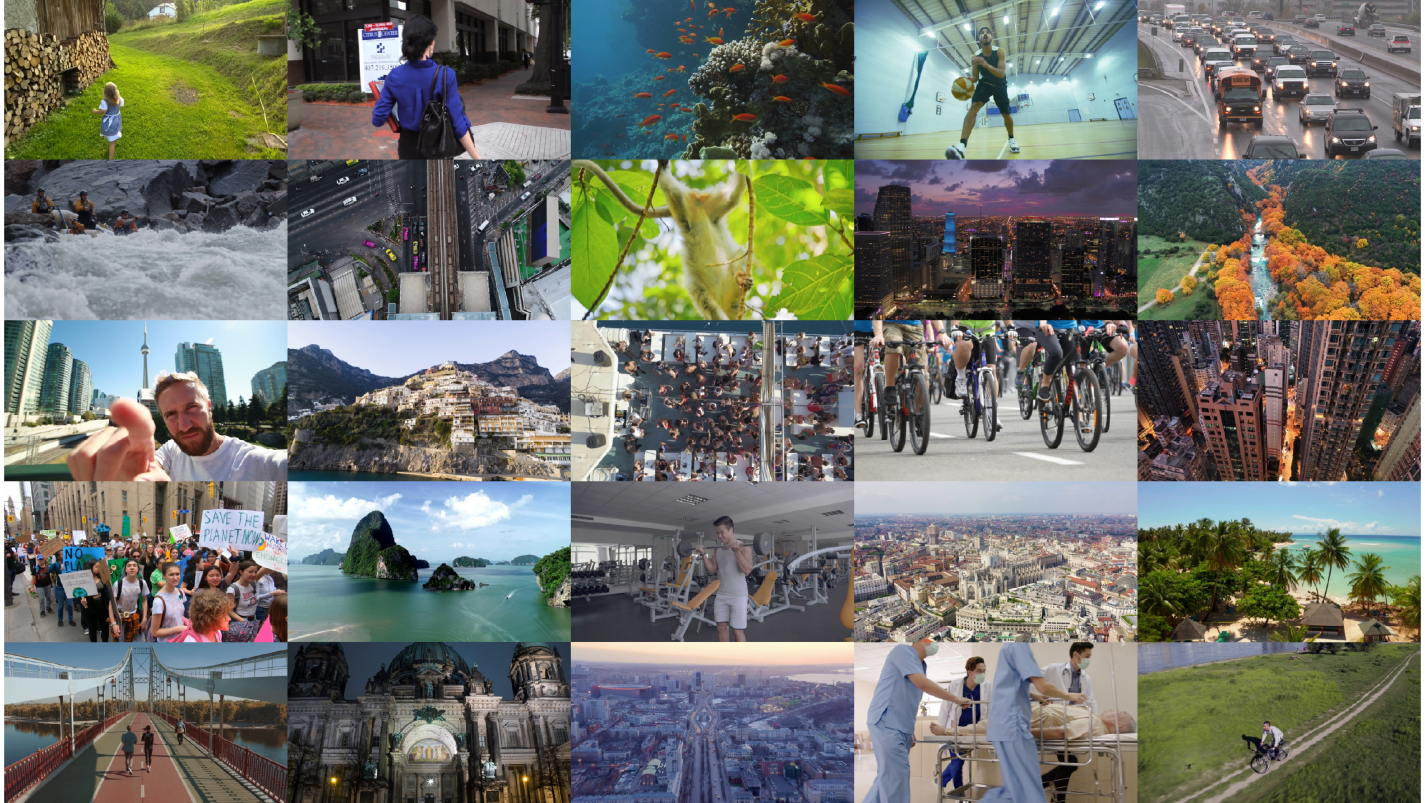}
    \caption{A snapshot from the POND25 dataset. Scenes exhibit richer composition compared to previous validation datasets.}
    \label{fig:supp_pond5_snapshot}
\end{figure*}

\begin{figure*}[ht]
    \centering
    \includegraphics[width=\linewidth]{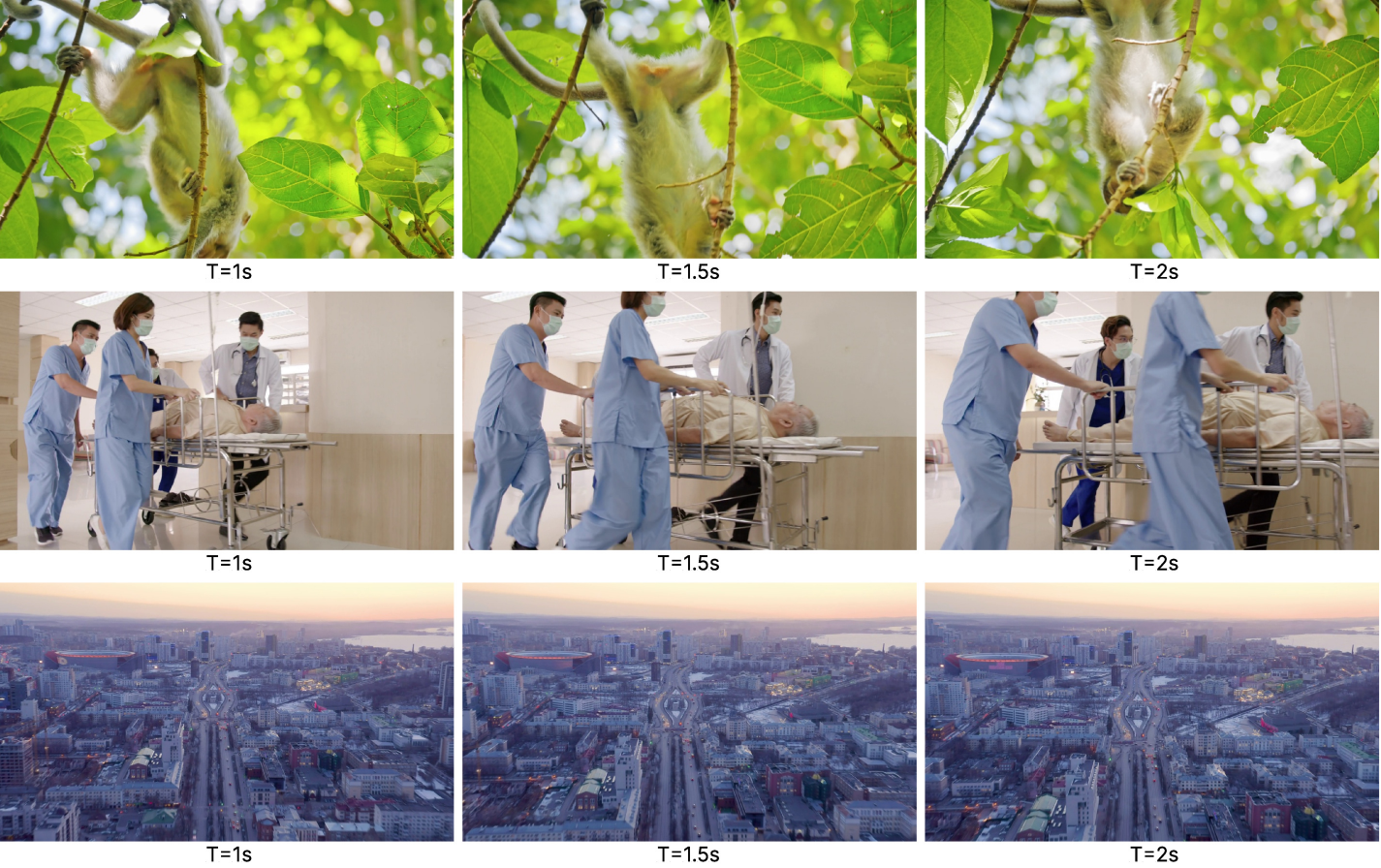}
    \vspace{-3mm}
    \caption{A snapshot from the POND25 dataset. Scenes exhibit higher motion complexity compared to previous validation datasets.}
    \label{fig:supp_pond5_temporal}
\end{figure*}

To evaluate model performance under complex and highly dynamic visual conditions, we construct a new validation dataset, denoted as \textbf{POND25}, derived from the POND5 video repository. The dataset consists of 25 video clips at 2K resolution and a frame rate of 24 FPS. Each clip is approximately four seconds long and contains no scene transitions. The dataset includes a diverse range of content, encompassing humans, animals, urban scenes, and natural landscapes.
As illustrated in Figure~\ref{fig:supp_pond5_snapshot} and Figure~\ref{fig:supp_pond5_temporal}, POND25 contains a variety of challenging cases such as aerial views, multi-subject compositions, and fast camera movements. Compared with earlier validation datasets (e.g., REDS40 and VideoLQ), which contain many handheld scenes, POND25 features only professionally filmed sequences with smooth camera motion and no shake or jitter.

\section{Implementation Details}

\subsection{Pseudo Code of Context-Parallel in HYPI2VR}

\begin{algorithm}[!b]
\resizebox{0.95\linewidth}{!}{
\begin{minipage}{\linewidth}
\caption{Context-Parallel Temporal Attention in HYPI2VR}
\label{alg:context_parallel}
\begin{algorithmic}[1]
\Require Hidden states $\mathbf{H}$, context-parallel group $\mathcal{G}_{cp}$
\State $\mathbf{Q} \gets \text{Attn.to\_q}(\mathbf{H})$
\State $\mathbf{K} \gets \text{Attn.to\_k}(\mathbf{H})$
\State $\mathbf{V} \gets \text{Attn.to\_v}(\mathbf{H})$
\State $\mathbf{Q} \gets \text{Attn.norm\_q}(\mathbf{Q})$
\State $\mathbf{K} \gets \text{Attn.norm\_k}(\mathbf{K})$
\vspace{0.5em}

\State \Comment{\textbf{Context-Parallel Synchronization}}
\State $\mathbf{K} \gets \textcolor{red}{\textbf{gather\_all\_from\_cp\_group}}(\mathbf{K}, \mathcal{G}_{cp})$
\State $\mathbf{V} \gets \textcolor{red}{\textbf{gather\_all\_from\_cp\_group}}(\mathbf{V}, \mathcal{G}_{cp})$
\vspace{0.5em}

\State $\mathbf{H} \gets \text{Attention}(\mathbf{Q}, \mathbf{K}, \mathbf{V})$
\State $\mathbf{H} \gets \text{Attn.to\_out}(\mathbf{H})$
\end{algorithmic}
\end{minipage}
}
\end{algorithm}

Since the 1D temporal attention block is the only module requiring inter-frame communication, \textbf{HYPI2VR} achieves context parallelism by synchronizing key and value tensors across devices within the temporal attention block. Algorithm~\ref{alg:context_parallel} illustrates the procedure.

\subsection{Inference Efficiency}
\label{sec:runtime}
We report the inference efficiency of HYPI2VR and prior methods in \tablename~\ref{tab:runtime}.
All experiments are conducted on a single NVIDIA H200 GPU for $\times4$ VSR using 50 frames of resolution $512\times512$.
While HYPI2VR is primarily designed to achieve significantly improved restoration quality, it maintains inference efficiency within the range of one-step models.
In particular, its runtime is comparable to SeedVR2-7B, despite not employing temporal compression, partly due to avoiding the computational overhead of 3D attention.
This indicates that the quality improvements of HYPI2VR come at a reasonable and practical computational cost.

\begin{table}[t]
\centering
\vspace{-2pt}
\caption{
Inference speed comparison on a single NVIDIA H200 GPU for $\times4$ VSR with 50 frames of resolution $512\times512$.
We report the number of sampling steps, runtime (ms), and model parameters.
}
\label{tab:runtime}
\vspace{2pt}
\begin{tabular}{lccc}
\toprule
Method & Steps & Runtime (ms) & Params \\
\midrule
UAV & 30 & 2080 & 1.4B \\
MGLD-VSR & 50 & 2369 & 1.4B \\
STAR & 15 & 1617 & 2.5B \\
SeedVR2-7B & 1 & 480 & 7B \\
DLoRAL & 1 & 198 & 1.3B \\
FlashVSR & 1 & 31 & 1.8B \\
HYPI2VR (Ours) & 1 & 513 & 13B \\
\bottomrule
\end{tabular}
\end{table}

\section{Extended Ablation and Analysis}
\label{sec:supp_ablation}
\subsection{Progressive Improvement of Temporal Consistency}

\begin{figure*}[t]
    \centering
    \includegraphics[width=\linewidth]{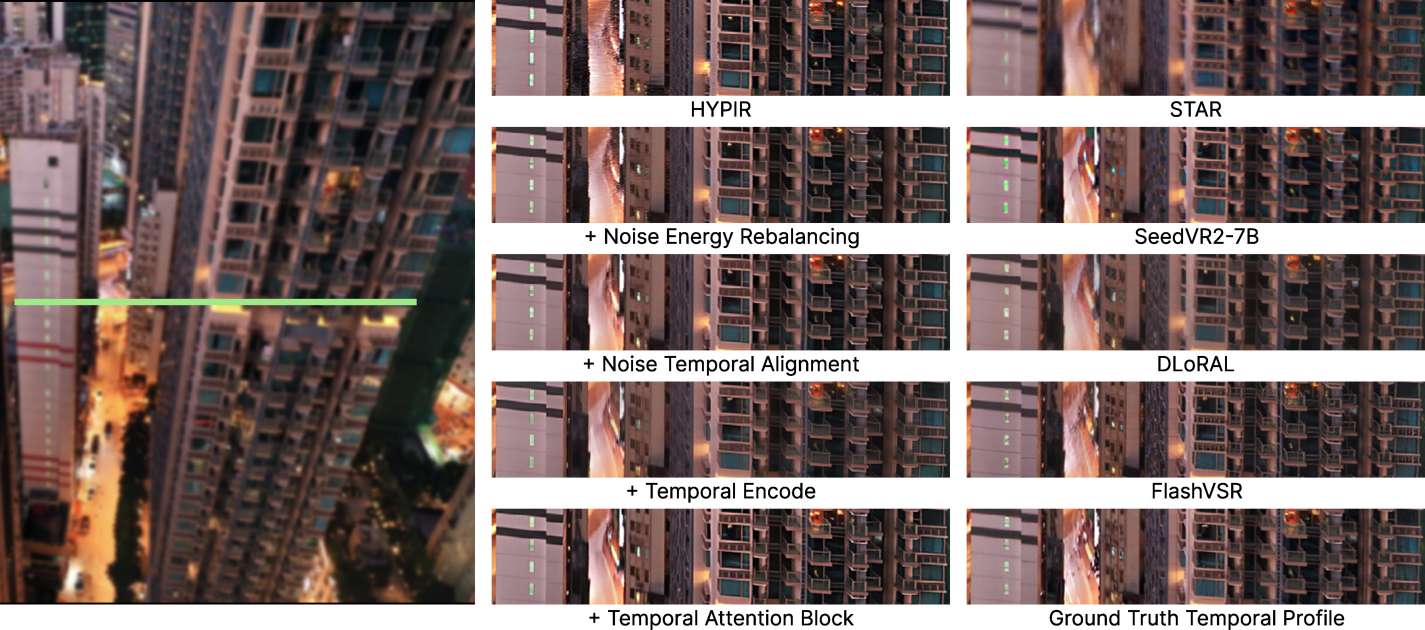}
    \caption{Temporal consistency profiles of different model variants in the ablation study on the POND25 dataset. The progressively added modules elevate the temporal consistency of HYPIR to a level comparable with other video restoration methods.}
    \label{fig:ablation_temp_profile}
\end{figure*}

As shown in Fig.~\ref{fig:ablation_temp_profile}, the proposed components progressively improve temporal consistency throughout restoration.
Starting from the HYPIR baseline, each additional module further stabilizes temporal variations, eventually achieving consistency comparable to dedicated video restoration methods.

\subsection{Temporal Modeling Design}

\paragraph{Temporal Attention Design.}

\begin{figure}[b]
\centering
\includegraphics[width=\linewidth]{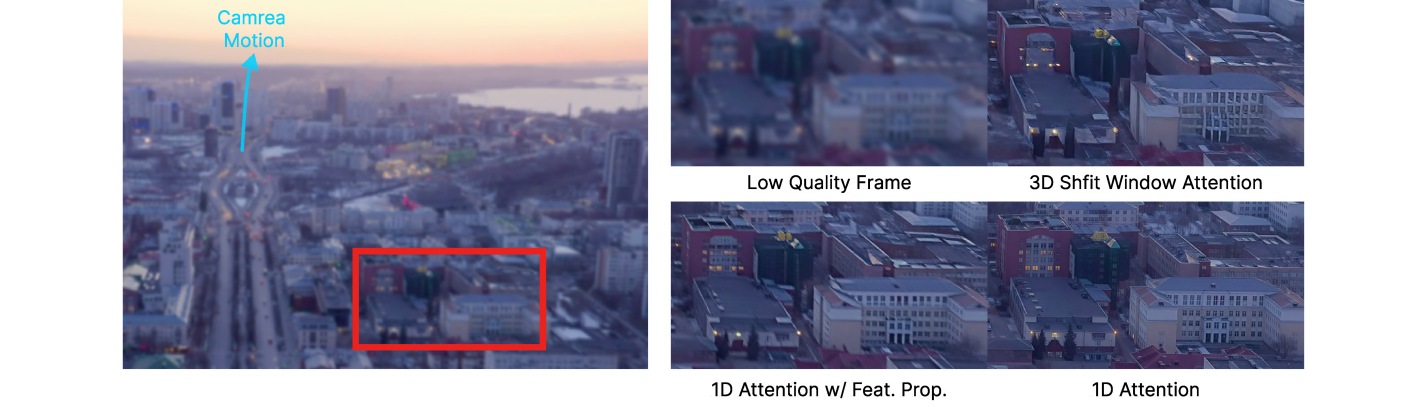}
\caption{Comparison of temporal modeling strategies under dynamic scenes with camera motion, where 3D attention introduces artifacts, feature propagation blurs details, and our method achieves a better balance between temporal consistency and detail preservation.}
\label{fig:ablation_dynamic}
\end{figure}

We further compare our lightweight 1D temporal attention with a 3D shift-window attention design following SeedVR2~\cite{seedvr2}.
As shown in Fig.~\ref{fig:ablation_dynamic}, replacing our temporal block with 3D shift-window attention leads to noticeably degraded restoration quality.
We attribute this to the higher optimization complexity of 3D attention, which jointly models spatial and temporal interactions and typically requires substantially larger training data and computation to converge effectively.
Under our training setup, decoupled 1D temporal attention is more stable and efficient, improving temporal consistency while preserving spatial detail synthesis of the image prior.

\paragraph{Effect of Feature Propagation.}
We also evaluate feature propagation, a common strategy for improving temporal consistency in prior T2I-based video restoration methods.
Although feature propagation can reduce frame-wise inconsistency, it relies heavily on optical-flow accuracy and may introduce propagation errors in regions with complex motion, occlusion, or fine structures.
As shown in Fig.~\ref{fig:ablation_dynamic}, this often degrades high-frequency details and weakens the visual richness provided by the image generative prior.
Therefore, we do not use feature propagation in the final model, and instead rely on noise alignment and temporal attention to improve consistency while preserving fine details.

\subsection{Hyper-Parameter Selection in Noise Re-balancing}

To determine an appropriate value of $\tau$, we train multiple HYPIR models from scratch with different $\tau$ settings.
Except for the noise rebalancing parameter, all training configurations are kept identical.
During inference, the same $\tau$ used in training is adopted for consistency.

\begin{figure}[b]
    \centering
    \includegraphics[width=0.8\linewidth]{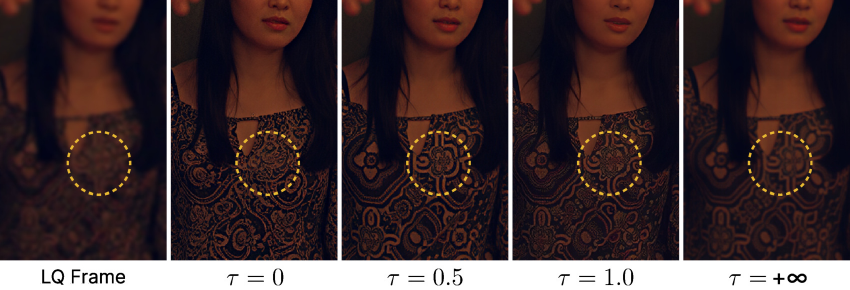}
    \caption{Visual comparison of restoration results with different $\tau$ values. 
Smaller $\tau$ introduces excessive low-frequency perturbations, while $\tau \rightarrow +\infty$ suppresses stochasticity and leads to overly smooth results. 
$\tau=1.0$ provides a better balance between structural fidelity and texture richness.}
    \label{fig:tau}
\end{figure}

As shown in \figurename~\ref{fig:tau} and \tablename~\ref{tab:tau}, different $\tau$ values lead to distinct restoration behaviors.
When $\tau=0$, low-frequency noise is largely preserved, resulting in degraded structural consistency.
As $\tau$ increases, low-frequency perturbations are progressively suppressed, improving visual fidelity and temporal coherence.
In the extreme case $\tau \rightarrow +\infty$, stochasticity is entirely removed, leading to overly smooth results with limited high-frequency details and reduced perceptual realism.
Compared with $\tau=0.5$, $\tau=1.0$ achieves a better balance between detail richness and structural consistency, and is thus adopted in all experiments.

\begin{table}[t]
\centering
\caption{
Quantitative evaluation under different $\tau$ values on POND25. 
$\tau=1.0$ achieves a favorable trade-off across fidelity, perceptual quality, and temporal consistency.
}
\label{tab:tau}
\vspace{2pt}
\begin{tabular}{c|cccc}
\toprule
\textbf{Metric} & $\tau=0$ & $\tau=0.5$ & $\tau=1.0$ & $\tau=+\infty$ \\
\midrule
PSNR$\uparrow$ & 23.90 & 25.09 & 25.80 & \textbf{28.30} \\
SSIM$\uparrow$ & 0.6826 & 0.7217 & 0.7428 & \textbf{0.8301} \\
LPIPS$\downarrow$ & 0.3421 & 0.2869 & 0.2663 & \textbf{0.2515} \\
MUSIQ$\uparrow$ & 61.80 & 60.76 & \textbf{62.08} & 37.88 \\
CLIP-IQA$\uparrow$ & \textbf{0.5035} & 0.4806 & 0.4729 & 0.3048 \\
DOVER$\uparrow$ & \textbf{0.6896} & 0.6571 & 0.6590 & 0.6143 \\
$E_{warp}\downarrow$ & 10.06 & 8.24 & 7.05 & \textbf{2.47} \\
\bottomrule
\end{tabular}
\end{table}

\subsection{Case Study on Motion Blur}
\begin{figure*}[t]
    \centering
    \includegraphics[width=\linewidth]{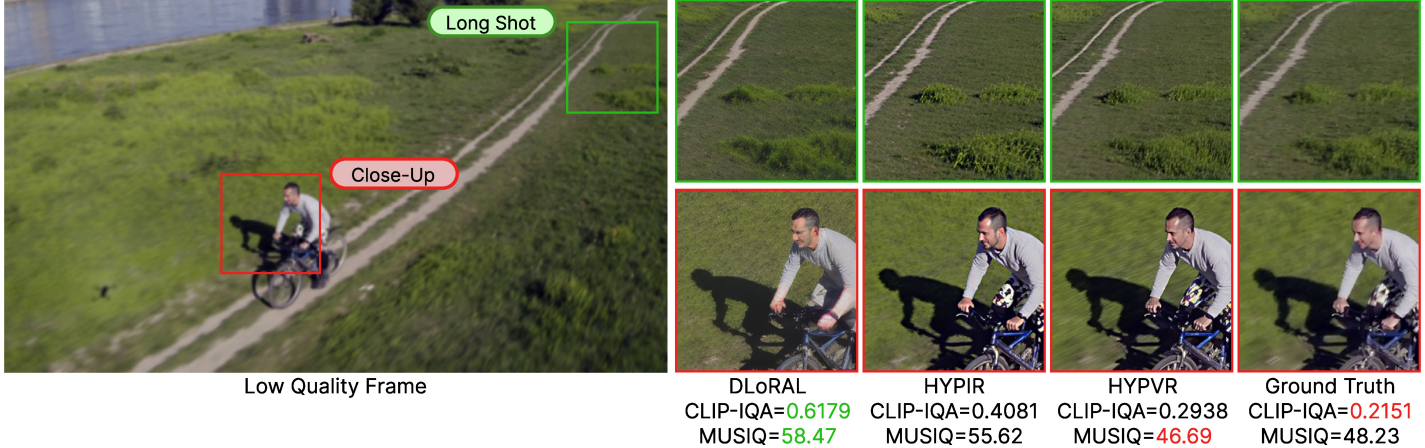}
    \caption{Effect of frame-based pre-training on dynamic scenes. Training with image frames allows the model to capture natural motion cues and reproduce realistic dynamic blur, outperforming models trained only on static images.}
    \label{fig:ablation_motion_blur}
\end{figure*}

As illustrated in \figurename~\ref{fig:ablation_motion_blur}, the observed improvements are primarily attributed to the stronger semantic and structural understanding enabled by our generative image model.
When applied to video restoration, HYPI2VR is able to reason about the presence of motion blur and its relationship to scene semantics, distinguishing naturally blurred regions from structures that should remain sharp.
This allows HYPI2VR to reconstruct videos with more realistic temporal behavior and natural variations in blur.
In contrast, methods lacking such semantic understanding tend to over-sharpen motion-blur regions and ignore natural motion patterns.
Although these methods may achieve higher scores on perceptual metrics such as CLIP-IQA, this is largely because such metrics emphasize global semantics and high-frequency content rather than the physical plausibility of motion and blur.

\section{More Results}
\label{sec:appendix_more_results}
We provide additional qualitative comparisons to further demonstrate the effectiveness and generalization of \textbf{HYPI2VR}.
These results include supplementary comparisons on the POND25 dataset, as well as additional real-world and historical film restoration examples.

\begin{figure*}
    \centering
    \makebox[\linewidth][c]{
        \includegraphics[width=1.1\linewidth]{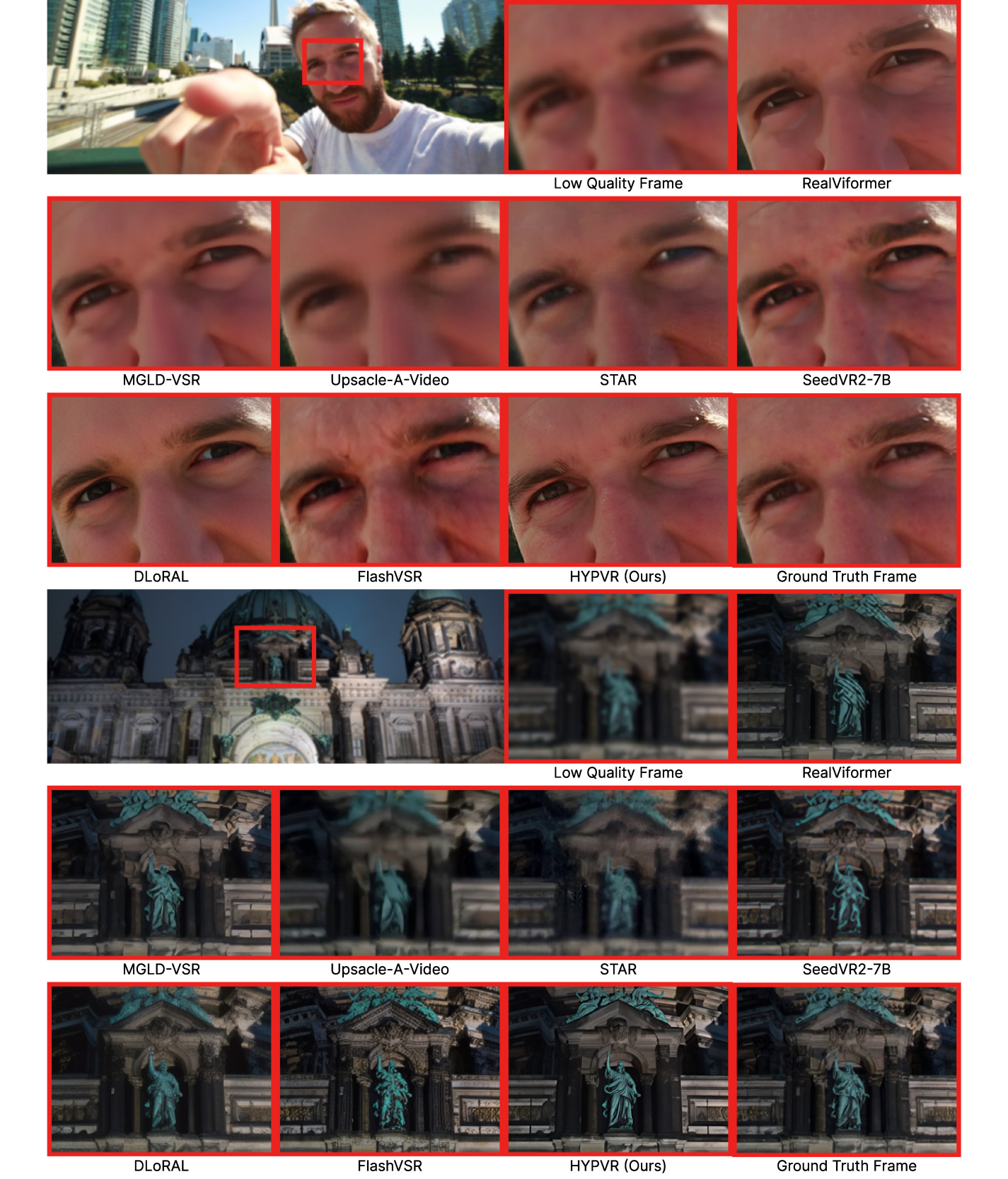}
    }
    \caption{Additional qualitative comparisons on the POND25 dataset. HYPI2VR accurately reconstructs complex spatial structures and reproduces realistic textures under dynamic motion.}
    \label{fig:supp_visual_syn_pond1}
\end{figure*}

\begin{figure*}
    \centering
    \makebox[\linewidth][c]{
        \includegraphics[width=1.1\linewidth]{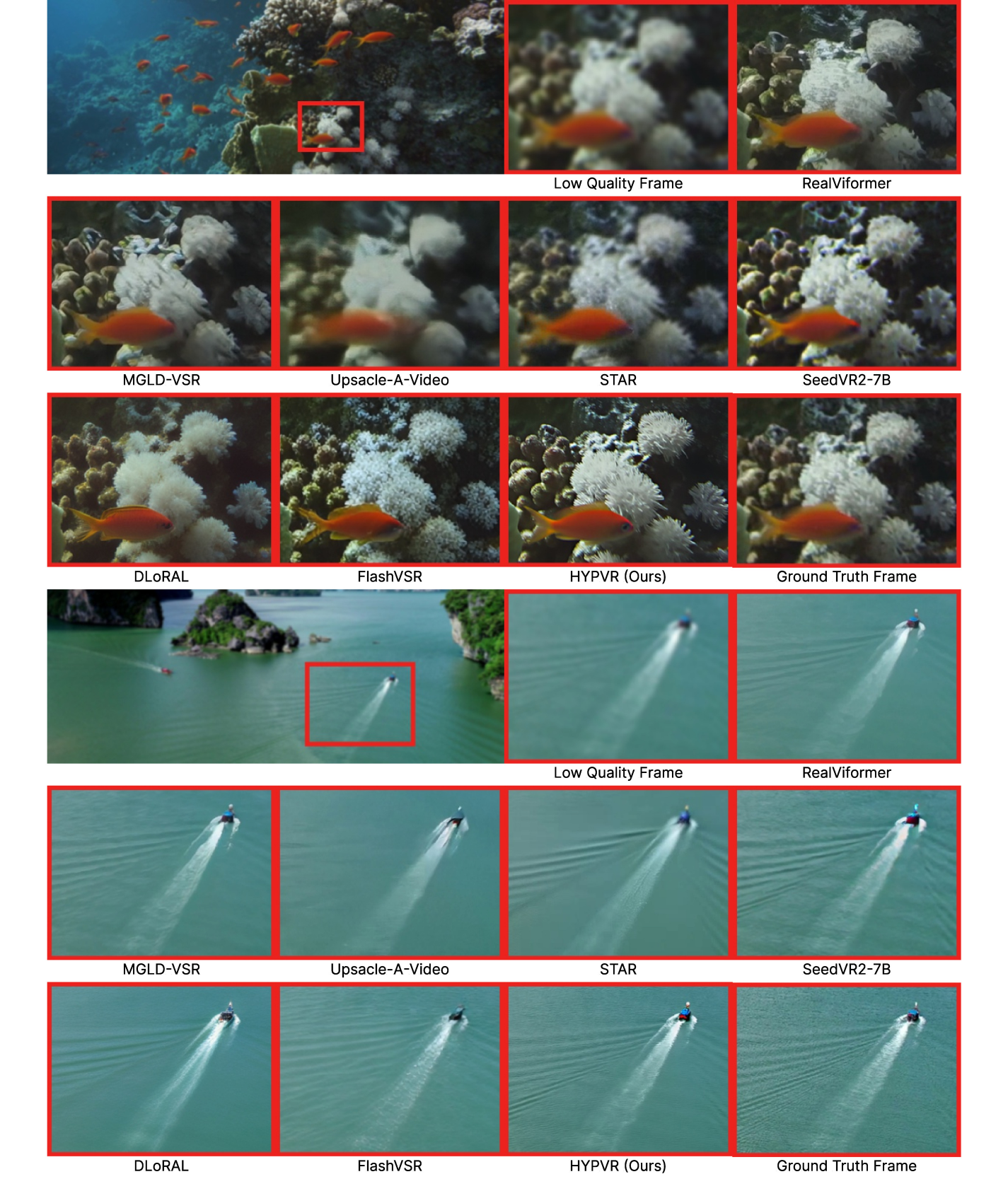}
    }
    \caption{Additional qualitative comparisons on the POND25 dataset. HYPI2VR accurately reconstructs complex spatial structures and reproduces realistic textures under dynamic motion.}
    \label{fig:supp_visual_syn_pond2}
\end{figure*}

\begin{figure*}
    \centering
    \makebox[\linewidth][c]{
        \includegraphics[width=1.1\linewidth]{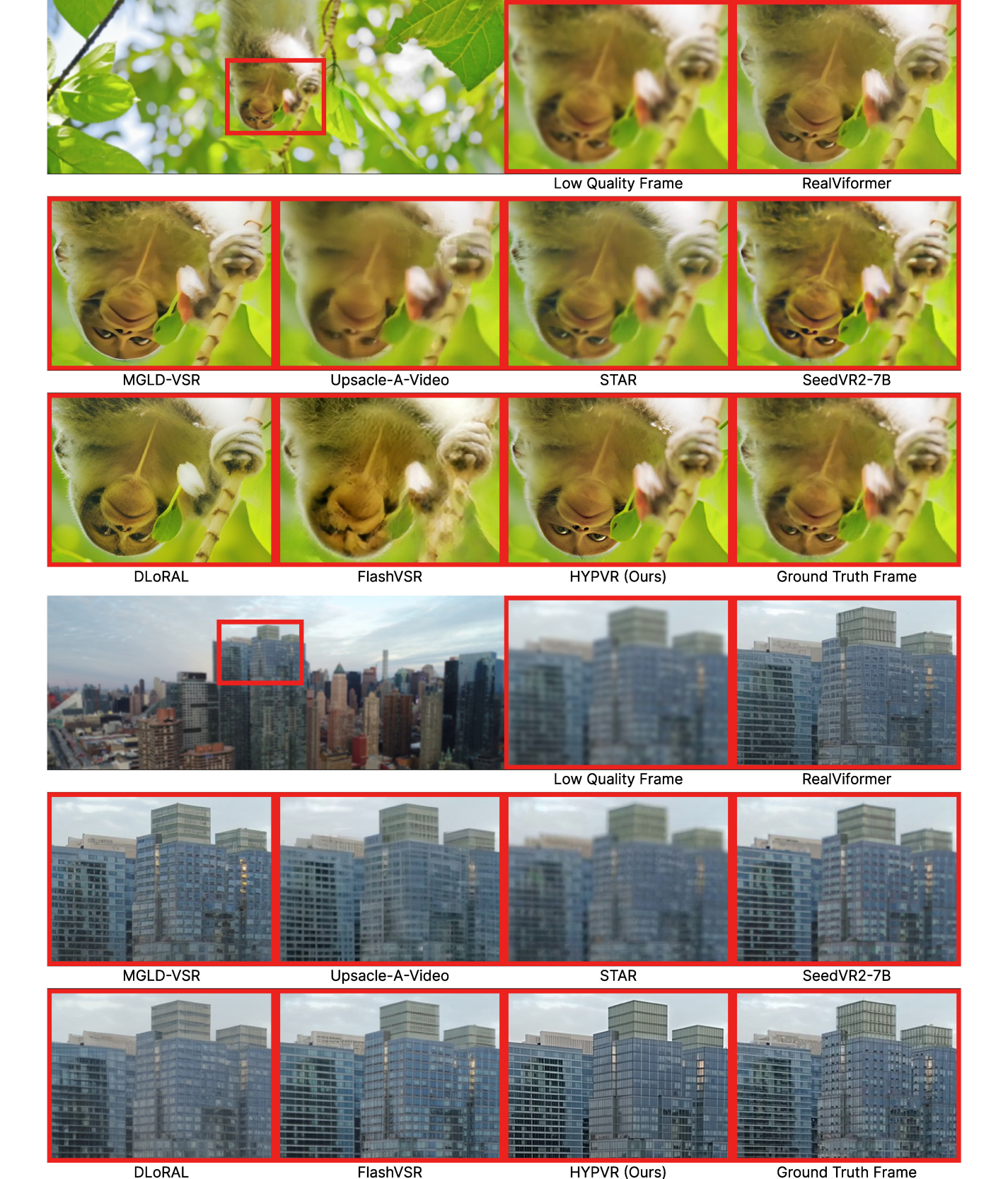}
    }
    \vspace{-5mm}
    \caption{Qualitative comparison on the POND25 dataset. HYPI2VR accurately reconstructs complex spatial structures and reproduces realistic textures under challenging dynamic motion.}
    \label{fig:supp_visual_syn_pond3}
\end{figure*}

\noindent\textbf{Additional Synthetic Video Results.}
We provide additional qualitative results on synthetic datasets to further demonstrate the effectiveness and generalization of \textbf{HYPI2VR}.
As shown in Figures~\ref{fig:supp_visual_syn_pond1}--\ref{fig:supp_visual_syn_pond3}, HYPI2VR consistently reconstructs fine spatial details and realistic textures even under complex motion.
The model performs particularly well in highly dynamic scenes, accurately interpreting depth and layer relationships while restoring motion-rich content with coherent structures and natural textures.
HYPI2VR also demonstrates strong spatial reasoning under complex geometries, such as aerial views of urban environments, precisely reconstructing fine architectural boundaries while preserving large-scale scene structure.

\begin{figure*}
    \centering
    \makebox[\linewidth][c]{
        \includegraphics[width=1.1\linewidth]{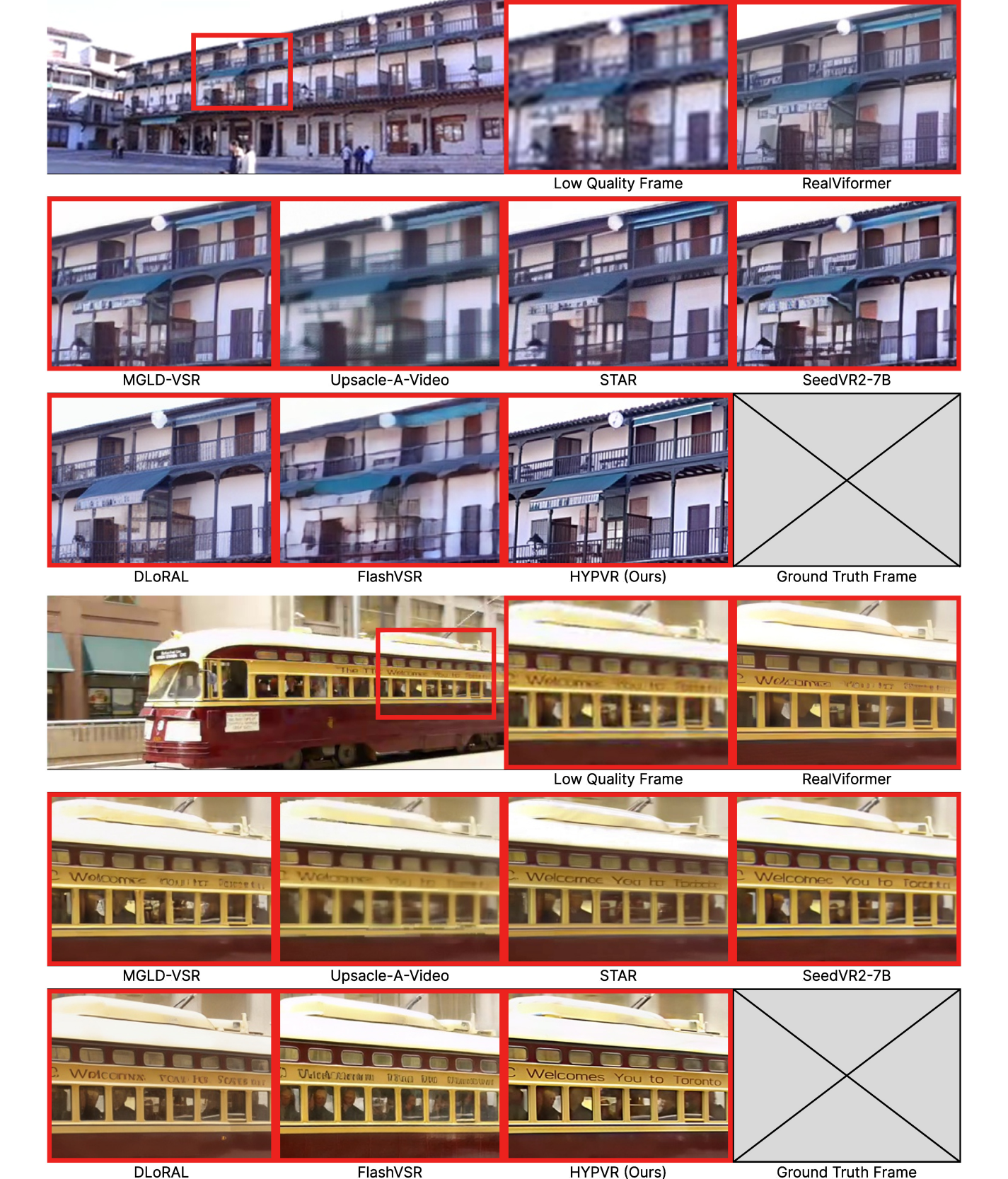}
    }
    \vspace{-5mm}
    \caption{Qualitative comparison on the real-world VideoLQ dataset. HYPI2VR preserves structure and avoids distortion under challenging degradations.
}
    \label{fig:main_visual_real_videolq}
\end{figure*}

\noindent\textbf{Real-World Video Results.}
On the real-world VideoLQ dataset, \figurename~\ref{fig:main_visual_real_videolq} shows that the proposed noise warping enables HYPI2VR to maintain consistent geometry even in challenging high-motion regions, effectively avoiding the distortions commonly observed in competing methods.
HYPI2VR also accurately captures scene depth and motion cues, producing temporally stable and spatially coherent video frames that faithfully reflect real-world structures and dynamics.

\begin{figure*}
    \centering
    \makebox[\linewidth][c]{
        \includegraphics[width=1.1\linewidth]{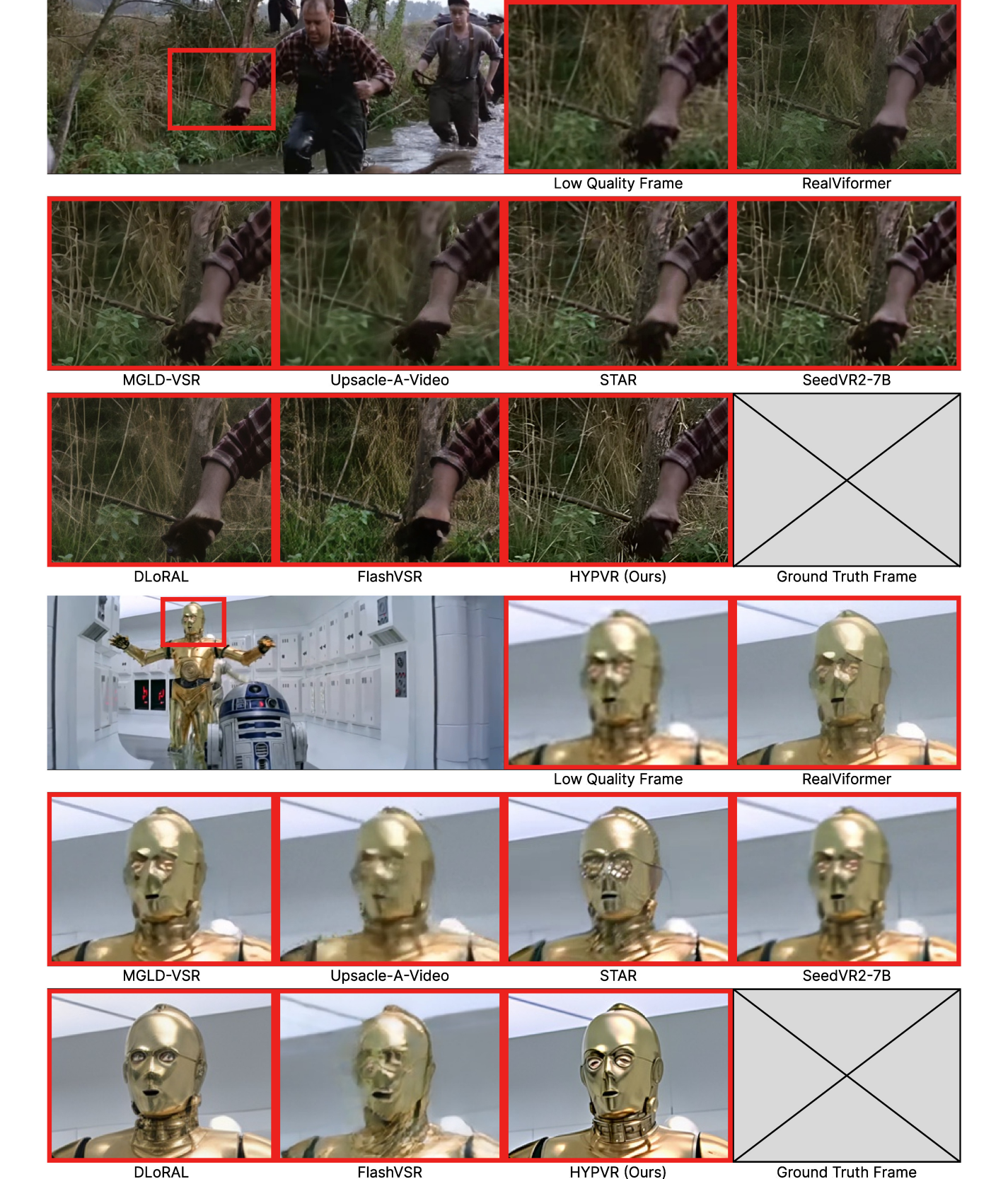}
    }
    \vspace{-5mm}
    \caption{Qualitative restoration on historical film footage. HYPI2VR recovers facial details and background textures while maintaining temporal coherence.}
    \label{fig:main_visual_real_film1}
\end{figure*}

\begin{figure*}
    \centering
    \makebox[\linewidth][c]{
        \includegraphics[width=1.1\linewidth]{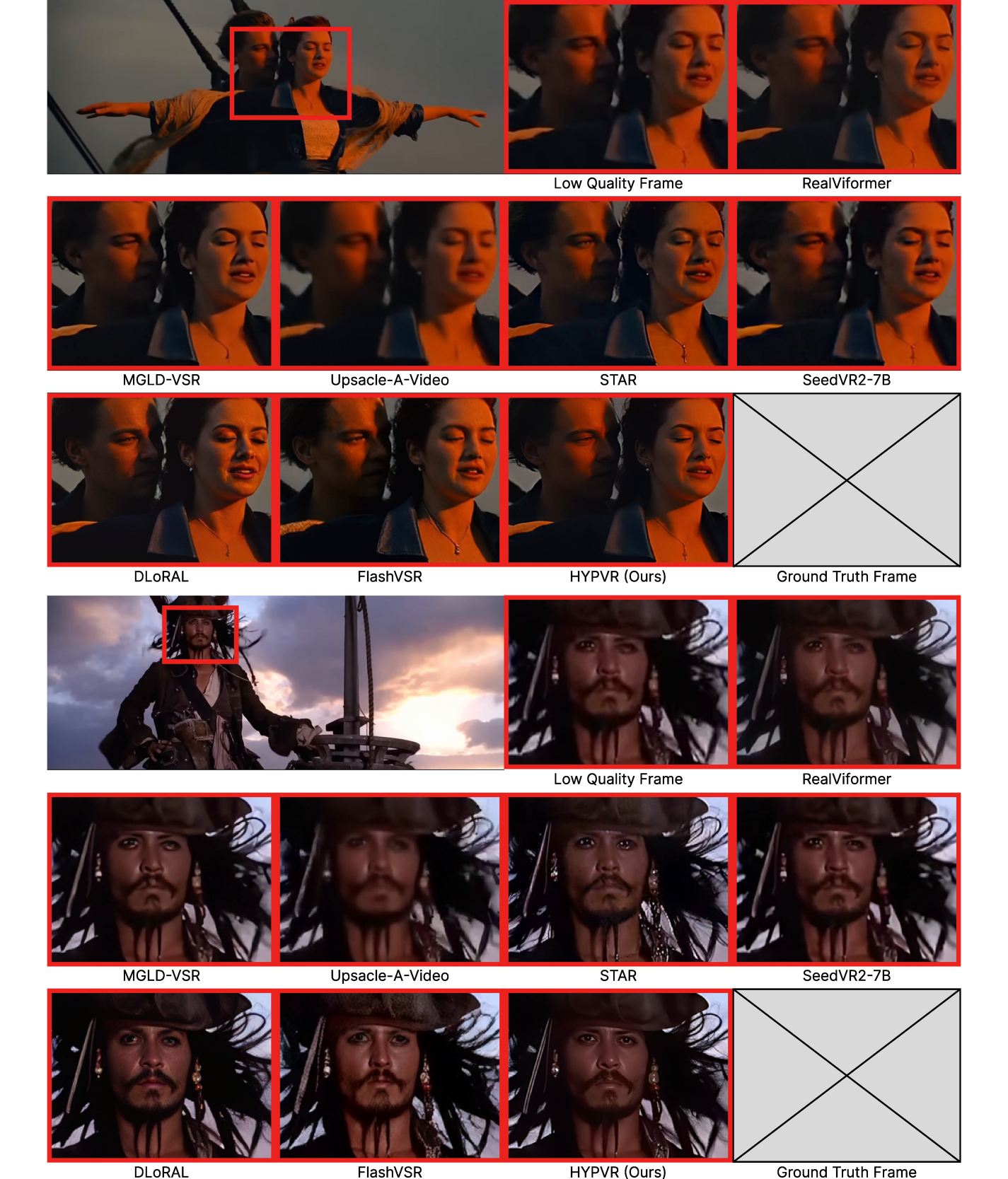}
    }
    \vspace{-5mm}
    \caption{Qualitative restoration on historical film footage. HYPI2VR reconstructs clear facial features and background details with stable visual quality.}
    \label{fig:main_visual_real_film2}
\end{figure*}

\noindent\textbf{Historical Film Restoration.}
\figurename~\ref{fig:main_visual_real_film1} and \figurename~\ref{fig:main_visual_real_film2} present qualitative comparisons of historical film footage.
Across a wide range of real-world degradations and content variations, HYPI2VR consistently generates visually coherent results with semantically faithful textures.
These examples highlight the strong generalization ability of our method and its capacity to synthesize perceptually natural details in complex real-world scenarios.


\clearpage

\end{document}